\documentclass[twoside]{article}

\usepackage[accepted]{aistats2025}
\usepackage[utf8]{inputenc} 
\usepackage[T1]{fontenc}    
\usepackage{hyperref}       
\usepackage{url}            
\usepackage{booktabs}       
\usepackage{amsfonts}       
\usepackage{nicefrac}       
\usepackage{microtype}      
\usepackage{xcolor}         

\usepackage[utf8]{inputenc} 
\usepackage[T1]{fontenc}    
\usepackage{hyperref}       
\usepackage{url}            
\usepackage{booktabs}       
\usepackage{amsfonts}       
\usepackage{nicefrac}       
\usepackage{microtype}      
\usepackage{xcolor}         
\usepackage{subcaption}
\usepackage{multirow}
\usepackage{makecell}

\usepackage{amsmath, amssymb}
\usepackage{graphicx}
\usepackage[font=normalsize,labelfont=bf]{caption}
\usepackage{tikz}
\usepackage{natbib}
\newtheorem{theorem}{Theorem}[section]
\newtheorem{lemma}[theorem]{Lemma}

\newtheorem{corollary}[theorem]{Corollary}
\newtheorem{definition}[theorem]{Definition}
\usetikzlibrary{positioning, arrows.meta, backgrounds, calc}
\usepackage{setspace}

\newcommand{\ind}{\perp\!\!\!\!\perp}

\newcommand{\myfootnote}[1]{%
  \refstepcounter{footnote}
  \hyperlink{fn:\thefootnote}{\textsuperscript{\thefootnote}}
  \footnotetext{\hypertarget{fn:\thefootnote}{}#1}
}

\begin{document}

%

%
\runningauthor{Shyam Venkatasubramanian, Ali Pezeshki, Vahid Tarokh}

\twocolumn[

\aistatstitle{Steinmetz Neural Networks for Complex-Valued Data}

\aistatsauthor{\href{mailto:<shyam.venkatasubramanian@duke.edu>?Subject=Steinmetz Neural Networks}{Shyam Venkatasubramanian} \And \href{mailto:<ali.pezeshki@colostate.edu>?Subject=Steinmetz Neural Networks}{Ali Pezeshki} \And  \href{mailto:<vahid.tarokh@duke.edu>?Subject=Steinmetz Neural Networks}{Vahid Tarokh} }

\aistatsaddress{ Duke University \And  Colorado State University \And Duke University} ]

\begin{abstract}
  We introduce a new approach to processing complex-valued data using DNNs consisting of parallel real-valued subnetworks with coupled outputs. Our proposed class of architectures, referred to as \textit{Steinmetz Neural Networks}, incorporates multi-view learning to construct more interpretable representations in the latent space. Moreover, we present the \textit{Analytic Neural Network}, which incorporates a consistency penalty that encourages analytic signal representations in the latent space of the Steinmetz neural network. This penalty enforces a deterministic and orthogonal relationship between the real and imaginary components. Using an information-theoretic construction, we demonstrate that the generalization gap upper bound posited by the analytic neural network is lower than that of the general class of Steinmetz neural networks. Our numerical experiments depict the improved performance and robustness to additive noise, afforded by our proposed networks on benchmark datasets and synthetic examples.
\end{abstract}

\section{Introduction}

In recent years, the advancement of neural networks has spurred a wealth of research into specialized models designed for processing complex-valued data. Complex-valued neural networks (CVNNs) are a pivotal area of focus due to their intrinsic capability to leverage both the magnitude information and the phase information embedded in complex-valued signals, offering a distinct advantage over their real-valued counterparts (RVNNs) \citep{hirose2012complex, guberman2016complex, trabelsi2017deep}. This is crucial across a spectrum of applications including telecommunications, medical imaging, and radar and sonar signal processing \citep{Virtue2017BetterTR, gao2019radar, smith2023complexvalued}. However, CVNNs are often encumbered by higher computational costs and more complex training dynamics \citep{bassey2021survey, lee2022survey, wu2023complex}. These difficulties arise from the necessity to manage and optimize parameters in the complex domain, which can lead to instability in gradient descent methods and challenges in network convergence. Additionally, the search for effective and efficient complex-valued activation functions is a challenge \citep{scardapane2020complex, lee2022survey}.

While much of the discussion comparing CVNNs and RVNNs has been focused on the theoretical aspects of CVNNs, the development of improved RVNN architectures for the processing of complex-valued data remains an open problem. We propose to address this problem through a feature learning perspective, leveraging multi-view representation fusion \citep{sun2013survey, lahat2015multimodal, zhao2017multi}. By considering the independent and joint information of real and imaginary parts in successive processing steps, we aim to better capture the task-relevant information in complex-valued data.

In response to these considerations, this paper introduces the \textit{Steinmetz Neural Network} architecture, a real-valued neural network that incorporates multi-view learning to improve the processing of complex-valued data for predictive tasks. This architecture aims to mitigate the challenges of training CVNNs while forming more interpretable latent space representations, and comprises separate subnetworks that independently filter the irrelevant information present within the real and imaginary components, followed by a joint processing step. We note that the task-relevant interactions between components are not lost during the separate processing step, as they are handled in joint processing.

An advantage afforded by the Steinmetz neural network's initial separate processing approach is that it provides control over the coherent combination of extracted features before joint processing. This choice is critical, as the proper combination of these features can lead to improved generalization. A key innovation from our approach is the derivation of a consistency constraint that encourages the extracted real and imaginary features to be related through a deterministic function, which lowers the Steinmetz neural network's generalization upper bound. For practical implementation, we choose this function to be the discrete Hilbert transform, since it ensures orthogonality between the extracted features to increase diversity \citep{cizek1970hilbert, chaudhry2020continual}. This methodology, referred to as the \textit{Analytic Neural Network}, attempts to leverage these structured representations to achieve improved generalization over the class of Steinmetz networks.

The organization of this paper is as follows. In Section \ref{sec:preliminaries}, we review complex and analytic signal representations, and survey the related work from CVNN literature. In Section \ref{sec:steinmetz_neural_networks}, we present the Steinmetz neural network architecture and discuss its theoretical foundations. In Section \ref{sec:consistency_constraint}, we summarize the consistency constraint and provide generalization gap bounds for the Steinmetz network. In Section \ref{sec:analytic_neural_network}, we introduce the analytic neural network and the Hilbert transform consistency penalty. In Section \ref{sec:empirical_results}, we present empirical results on benchmark datasets for complex-valued multi-class classification and regression, and provide synthetic experiments. In Section \ref{sec:conclusion}, we summarize our work. 

Our main contributions are summarized as follows: 
\begin{enumerate}
\item We introduce the \textit{Steinmetz Neural Network}, a real-valued neural network architecture that leverages multi-view learning to construct more interpretable latent space representations.
\item We propose a consistency constraint on the latent space of the Steinmetz neural network to obtain a smaller upper bound on the generalization gap, and derive these generalization bounds.
\item We outline a practical implementation of this consistency constraint through the Hilbert transform, and present the \textit{Analytic Neural Network}, which promotes analytic signal representations within the latent space of the Steinmetz neural network.
\end{enumerate}

\section{Preliminaries} \label{sec:preliminaries}
To motivate our framework, we begin by formally defining complex and analytic signal representations, and review existing real-valued neural networks and complex-valued neural networks for complex signal processing. Let $\smash{\mathcal{U} = \{0,1,\ldots,N-1\}}$ and $\smash{\mathcal{V} = \{0,1,\ldots,M-1\}}$, where $N \in \mathbb{N}$ is the signal period and $M \in \mathbb{N}$ denotes the size of the training dataset, $s$. To characterize the uncertainty of these signal representations, we denote $X \in \mathbb{C}^{dN}$, $X_R, X_I \in \mathbb{R}^{dN}$ as the features, $Y \in \mathbb{C}^{k}$, $Y_R, Y_I \in \mathbb{R}^{k}$ as the labels, and $Z \in \mathbb{C}^{lN}$, $Z_R, Z_I \in \mathbb{R}^{lN}$ as the latent variables with $R,I$ denoting the respective real and imaginary parts, where:
\begin{alignat*}{6}
    &X = (X[0],..,X[N{-}1]), \, X_R = (X_R[0],...,X_R[N{-}1]), \\
    &X_I = (X_I[0],..,X_I[N{-}1]), \, Z = (Z[0],..,Z[N{-}1]), \\
    &Z_R = (Z_R[0],..,Z_R[N{-}1]), \, Z_I = (Z_I[0],..,Z_I[N{-}1]).
\end{alignat*}
Correspondingly, we denote the training dataset using $s = \{(x^m,y^m), \, m \in \mathcal{V}\}$. Let $P$ denote the joint probability distribution of $(X,Y)$, where $(X,Y) \sim P$, and suppose $\smash{(x^m,y^m) \overset{\text{i.i.d.}}\sim P}$. The product measure, $P^{\otimes M \vphantom{\prod}}$, posits $S = ((X^0,Y^0), (X^1,Y^1),\ldots,(X^{M-1},Y^{M-1}))$, where $S \sim P^{\otimes M \vphantom{\prod}}$, $s \sim P^{\otimes M \vphantom{\prod}}$, and $\smash{(X^m, Y^m) \overset{d}= (X, Y)}$.
 
\subsection{Complex Signal Representation}
Consider the complex stochastic process $\{X[n], n \in \mathcal{U}\}$, which comprises the individual real stochastic processes $\{X_R[n], n \in \mathcal{U}\}$ and $\{X_I[n], n \in \mathcal{U}\}$, wherein $X[n] = X_R[n] + iX_I[n]$, $\forall n \in \mathcal{U}$, with $X[n] \in \mathbb{C}^d$. The respective realizations of $\{X_R[n], n \in \mathcal{U}\}$ and $\{X_I[n], n \in \mathcal{U}\}$, denoted by $\{x_R[n],n \in \mathcal{U}\}$ and $\{x_I[n], n \in \mathcal{U}\}$, are real signals. These realizations define the complex signal $\{x[n],n \in \mathcal{U}\}$, wherein $x[n] = x_R[n] + ix_I[n]$, $\forall n \in \mathcal{U}$. This approach is rooted in the foundational work on the complex representation of AC signals \citep{Steinmetz1893}. The transformation of $X^m$ into $Y^m$ is given by:
\begin{align}
\begin{split}
    Y^m &= Y_R^m + iY_I^m \\
    &= \nu(X_R^m, X_I^m) + i\omega(X_R^m, X_I^m) \\
    &= \xi(X_R^m + iX_I^m) = \xi(X^m)
\end{split}
\end{align}

\noindent This transformation showcases the method by which complex signal representations, governed by their underlying stochastic properties, are processed for predictive tasks. The true function, $\xi(\cdot)$, which comprises the real-valued functions, $\nu(\cdot)$ and $\omega(\cdot)$, acts on random variables $X_R^m$ and $X_I^m$ to yield $Y_R^m$ and $Y_I^m$. Analytic signals are an extension of this framework and have no negative frequency components.

\subsubsection{Analytic Signal Representation} \label{sec:analytic_signals}
The analytic signal representation is defined as $X_I^m = \mathcal{H}\{ X_R^m\}$, where for $x_R^m \sim X_R^m$, we have $x_I^m = \mathcal{H}\{x_R^m\} \in \mathbb{R}^{dN}$, $\forall m \in \mathcal{V}$. This construction is formalized as:
\begin{align}
    &X^m = X_R^m + i\, \mathcal{H}\{X_R^m\}, \quad \text{where:}\\
    &\mathcal{H}\{X_R^m\}[n] = \frac{2}{dN} \sum_{u \in \mathcal{U}'} X_R^m[u]\, \cot\!\left[(u - n)\frac{\pi}{dN}\right].
\end{align}
$\mathcal{H}\{X_R^m\}$ is the discrete Hilbert transform (DHT) of $X_R^m$, where $\mathcal{U}' = \{\, u \in \{0,1,\dots,dN-1\}, u \not\equiv n \pmod{2} \,\}$, which selects even indices when $n$ is odd and odd indices when $n$ is even. The DHT introduces a phase shift of $-90^{\circ}$ to all the positive frequency components of $X_R^m$ and $+90^{\circ}$ to all the negative frequency components of $X_R^m$, establishing orthogonality between $X_R^m$ and $X_I^m$. We observe that $\mathcal{H}\{\cdot\}$ is bijective since it is invertible, wherein $-\mathcal{H}\{\mathcal{H}\{X_R^m\}\} = X_R^m$, $\forall m \in \mathcal{V}$.

\subsection{Related Work}
The development of neural networks for complex signal processing has led to extensive research comparing the effectiveness of complex-valued neural networks (CVNNs) versus real-valued neural networks (RVNNs). While CVNNs are theoretically capable of capturing the information contained in phase components --- see radar imaging \citep{gao2019radar}, electromagnetic inverse scattering \citep{guo2021electromagnetic}, MRI fingerprinting \citep{Virtue2017BetterTR}, and automatic speech recognition \citep{ShafranBS18} --- practical implementation is yet to show considerable improvements over RVNNs. In particular, \citep{guberman2016complex} depicts that RVNNs tend to have significantly lower training losses compared to CVNNs, and that comparing test losses, RVNNs still marginally outperform CVNNs in terms of generalization despite their vulnerability to overfitting. When the overfitting is substantial \citep{barrachina2021cvnn}, RVNNs, despite their simpler training and optimization, observe worse generalization than CVNNs. A similar result is illustrated in \citep{trabelsi2017deep}, where multidimensional RVNNs, with concatenated real and imaginary components fed into individual channels, exhibit performance metrics closely aligned with those of CVNNs, especially in architectures with constrained parameter sizes. Furthermore, CVNNs introduce higher computational complexity and encounter challenges in formulating holomorphic activation functions \citep{lee2022survey}. These studies reveal a balance between the theoretical benefits of CVNNs and the practical efficiencies of RVNNs. While CVNNs have received tremendous attention in recent years, the optimization of RVNNs for complex signal processing remains an open problem, especially in regard to improved training and regularization techniques for enhanced generalization performance and latent space interpretability.

\section{Steinmetz Neural Networks} \label{sec:steinmetz_neural_networks}
Reflecting on the inherent challenges and benefits of both RVNNs and CVNNs, we target a framework that leverages the simplicity in training offered by RVNNs while offering improved generalization in the processing of complex signals. Within this context, multi-view representation fusion emerges as a potential framework, proposing that different perspectives --- or `views' --- of data can provide complementary information, thereby enhancing learning and generalization \citep{sun2013survey, xu2013survey, lahat2015multimodal, zhao2017multi, yan2021deep}. Informed by this principle, we introduce the \textit{Steinmetz Neural Network} architecture, which is designed to process the real ($X_R^m$) and the imaginary ($X_I^m$) parts of complex signal representations as separate views before joint processing. We formalize how this architecture leverages the complementarity principle of multi-view learning in Section \ref{sec:steinmetz_theory}.

\subsection{Theoretical Foundations} \label{sec:steinmetz_theory}
\begin{figure*}[t!]
\centering
\begin{tikzpicture}[node distance=.3cm and 1.2cm, auto, thick, >=Latex, scale=0.8, every node/.style={scale=0.8}]
  \node (Y) {\(Y^m\)};
  \node (X) [right=of Y] {\(X^m\)};
  \node (Z) [right=of X] {\(Z^m\)};
  \node (Yhat) [right=of Z] {\(\hat{Y}^m\)};

  \node (XR) [right=of Yhat] {\((X_R^m, X_I^m)\)};
  \node (ZR) [right=of XR] {\((Z_R^m, Z_I^m)\)};
  \node (YhatR) [right=of ZR] {\(\hat{Y}^m\)};

  \draw[->] (Y) -- (X) node[midway, above] {\(\xi^{-1}(\cdot)\)};
  \draw[->] (X) -- (Z) node[midway, above] {\(\psi(\cdot)\)};
  \draw[->] (Z) -- (Yhat) node[midway, above] {\(h(\cdot)\)};

  \draw[->] (XR) -- (ZR) node[midway, above] {\(\psi^*(\cdot)\)};
  \draw[->] (ZR) -- (YhatR) node[midway, above] {\(h^*(\cdot)\)};

  \begin{scope}[on background layer]
    \draw[color=red!100, dotted, thick] ($(X.north west)+(0.05,0.4)$)  rectangle ($(Yhat.south east)+(0,-0.2)$);
    \draw[color=red!100, dotted, thick] ($(XR.north west)+(0.05,0.4)$) rectangle ($(YhatR.south east)+(0,-0.2)$);
  \end{scope}

\end{tikzpicture}
\caption{Classical RVNN Markov chain (left) and practical implementation (right) } \label{fig:markov_NN}
\end{figure*}
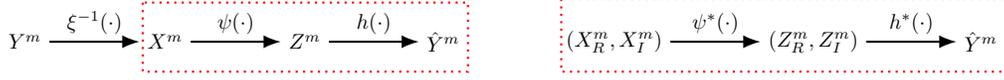

In the context of neural networks' information-theoretic foundations, \citep{Tishby2015} leveraged the Data Processing Inequality to characterize the architecture illustrated in Figure~\ref{fig:markov_NN}. This architecture aligns with the classical RVNN architecture from the literature on CVNNs \citep{trabelsi2017deep}. Here, $\xi(\cdot)$ denotes the true function, $h(\psi(\cdot))$ describes the neural network, and $\smash{\hat{Y}}^m = h(\psi(X^m))$ denotes the predictions. Accordingly, we have that $I(Y^m ; X^m) \geq I(Y^m ; Z^m) \geq I(Y^m ; \smash{\hat{Y}}^m)$. Regarding practical implementation, the input space, $\smash{[[X_R^m]^T, [X_I^m]^T]^T \in \mathbb{R}^{2dN}}$, is jointly processed by $\psi^*(\cdot)$, using individual channels for $X_R^m$ and $X_I^m$ to form the latent space, $\smash{[[Z_R^m]^T, [Z_I^m]^T]^T \in \mathbb{R}^{2lN}}$, which then gets jointly processed by $h^*(\cdot)$, using individual channels for $Z_R^m$ and $Z_I^m$, to obtain the output space, $\smash{\hat{Y}^m \in \mathbb{C}^{k}}$.

\begin{figure*}[t!]
\centering
\begin{tikzpicture}[node distance=.3cm and 1.2cm, auto, thick, >=Latex, scale=0.8, every node/.style={scale=0.8}]
  \node (Y) {\(Y^m\)};
  \node (X) [right=of Y] {\(X^m\)};
  \node (Xreal) [above right=of X] {\(X_R^m\)};
  \node (Ximag) [below right=of X] {\(X_I^m\)};
  \node (Xtilimag) [right=of Ximag] {\(Z_I^m\)};
  \node (Xtilreal) [right=of Xreal] {\(Z_R^m\)};
  \node (Z) [right=4.8cm of X] {\(Z^m\)};
  \node (Yhat) [right=of Z] {\(\hat{Y}^m\)};

  \node (ZR) [right=0.65cm of Yhat] {\((Z_R^m, Z_I^m)\)};
  \node (YhatR) [right=of ZR] {\(\hat{Y}^m\)};

  \draw[->] (Y) -- (X) node[midway] {\(\xi^{-1}(\cdot)\)};
  \draw[->] (X) -- (Ximag) node[midway, above left] {};
  \draw[->] (X) -- (Xreal) node[midway, below left] {};
  \draw[->] (Ximag) -- (Xtilimag) node[midway] {\(f(\cdot)\)};
  \draw[->] (Xreal) -- (Xtilreal) node[midway] {\(g(\cdot)\)};
  \draw[->] (Xtilimag) -- (Z) node[midway, above] {};
  \draw[->] (Xtilreal) -- (Z) node[midway, below] {};
  \draw[->] (Z) -- (Yhat) node[midway, above] {\(h(\cdot)\)};
  
  \draw[->] (ZR) -- (YhatR) node[midway, above] {\(h^*(\cdot)\)};

  \begin{scope}[on background layer]
    \draw[color=red!100, dotted, thick] ($(Z.north west)+(0.05,0.4)$)  rectangle ($(Yhat.south east)+(0,-0.2)$);
    \draw[color=red!100, dotted, thick] ($(ZR.north west)+(0.05,0.4)$) rectangle ($(YhatR.south east)+(0,-0.2)$);
  \end{scope}

\end{tikzpicture}
\caption{Steinmetz neural network Markov chain (left) and practical implementation (right)} \label{fig:markov_steinmetz}
\end{figure*}
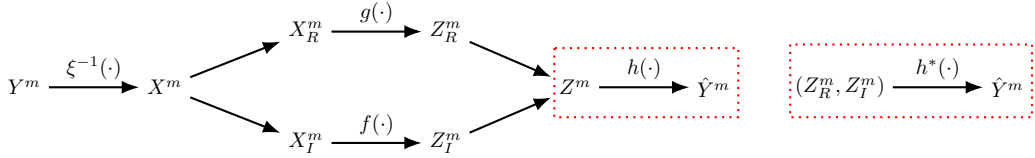

\noindent Building upon this architecture, the Steinmetz neural network postulates the Markov chain depicted in Figure~\ref{fig:markov_steinmetz}, where the random variables, $Z_I^m = f(X_I^m)$ and $Z_R^m = g(Z_R^m)$, are the respective outputs of the parallel subnetworks, $f(\cdot)$ and $g(\cdot)$, where $Z^m = Z_R^m + iZ_I^m$ denotes the latent representation, and $\smash{\hat{Y}}^m = h(Z^m)$ denotes the predictions yielded by the shared network, $h(\cdot)$ \myfootnote{For practical implementation, we mean center $Z_R^m$ and $Z_I^m$ as the final step before concatenation.}. Per the construction in Figure~\ref{fig:markov_steinmetz}, the Steinmetz neural network is given by:
\begin{align} \label{eq:steinmetz_architecture}
\begin{split}
    \hat{Y}^m &= h(Z_R^m + i Z_I^m) \\
    &= h(g(X_R^m) + i f(X_I^m)) = h(\psi(X^m)).
\end{split}
\end{align}

This architecture exhibits several distinctions from the classical RVNN. For practical implementation, $X_R^m$ and $X_I^m$ are processed by two different neural networks to form $Z_R^m$ and $Z_I^m$, respectively, which are concatenated in the latent space to form $\smash{[[Z_R^m]^T, [Z_I^m]^T]^T \in \mathbb{R}^{2lN}}$, and jointly processed by $h^*(\cdot)$ using individual channels for $Z_R^m$ and $Z_I^m$. The rationale behind this initial separate processing step stems from the complementarity principle of multi-view learning \citep{xu2013survey}. This principle suggests that before forming a shared latent space in a multi-view setting, separately processing views that contain unique information can improve the interpretability of representations. We now define relevant terms to extend this notion to our Steinmetz neural network architecture.

Suppose $X_R^m = (Z_R^m, \Lambda_R^m)$, $Z_R^m \ind \Lambda_R^m$, where $Z_R^m$ is the latent representation from $X_R^m$ that contains information relevant to $Y^m$ when combined with $Z_I^m$, and $\Lambda_R^m$ is information in $X_R^m$ irrelevant to $Y^m$ when combined with $Z_I^m$. Let $X_I^m = (Z_I^m, \Lambda_I^m)$, $Z_I^m \ind \Lambda_I^m$, wherein $Z_I^m$ is the latent representation from $X_I^m$ that contains information relevant to $Y^m$ when combined with $Z_R^m$, and $\Lambda_I^m$ is information in $X_I^m$ irrelevant to $Y^m$ when combined with $Z_I^m$. Let $(Z_R^m, Z_I^m) = (\smash{\hat{Y}}^m, \Gamma^m)$, $\smash{\hat{Y}}^m \ind \Gamma^m$, where $(Z_R^m, Z_I^m)$ is a sufficient statistic of $(X_R^m, X_I^m)$ with respect to $Y^m$. Here, $\smash{\hat{Y}}^m$ is a minimal sufficient statistic of $(Z_R^m, Z_I^m)$ with respect to $Y^m$, and $\Gamma^m$ is information in $(Z_R^m, Z_I^m)$ irrelevant to $Y^m$.

For the classical RVNN, we jointly process $(X_R^m,X_I^m)$ and aim to form $(Z_R^m, Z_I^m) = \psi^*(X_R^m, X_I^m)$, filtering out $(\Lambda_R^m, \Lambda_I^m)$, where $(Z_R^m, Z_I^m)$, is a sufficient statistic of $(X_R^m, X_I^m)$ with respect to $Y^m$. We then jointly process $(Z_R^m,Z_I^m)$ and attempt to form the predictions, $\smash{\hat{Y}}^m = h^*(Z_R^m, Z_I^m)$, filtering out $\Gamma^m$. We refer to this approach as \textit{joint-only processing}. For the Steinmetz neural network, we separately process $X_R^m$ and $X_I^m$, and attempt to form $Z_R^m = g(X_R^m)$ and $Z_I^m = f(X_I^m)$, filtering out $\Lambda_R^m$ and $\Lambda_I^m$, where $(Z_R^m, Z_I^m)$, is a sufficient statistic of $(X_R^m, X_I^m)$ with respect to $Y^m$. Paralleling the joint-only processing case, we jointly process this latent space, aiming to form $\smash{\hat{Y}}^m = h^*(Z_R^m, Z_I^m)$. We refer to this approach as \textit{separate-then-joint processing}.

In comparing the Steinmetz neural network's separate-then-joint processing scheme with the more classical joint-only processing scheme, we note that the former approach enables us to train individual subnetworks in place of $g(\cdot)$ and $f(\cdot)$ to filter out the respective information irrelevant to $Y^m$ present within $X_R^m$ and $X_I^m$. Contrarily, training a single neural network, $\psi^*(\cdot)$, via the joint-only processing scheme requires handling $X_R^m$ and $X_I^m$ simultaneously, meaning the network must learn to generalize across potentially disparate noise distributions and data characteristics. This makes it challenging to optimize the filtering of $\Lambda_R^m$ and $\Lambda_I^m$ if their properties differ significantly.

Accordingly, to characterize the complexity of representing $(Z_R^m, Z_I^m)$ from $(X_R^m, X_I^m)$ in the Steinmetz neural network and classical RVNN approaches, we propose the following construction. Let $\mathbf{\Sigma_J}$ denote the matrix of covariances of $X_R^m$, $X_I^m$ posited by the joint-only processing approach, and let $\mathbf{\Sigma_S}$ denote the matrix of covariances of $X_R^m$, $X_I^m$ posited by the separate-then-joint processing approach, where:
\begin{align}
   \mathbf{\Sigma_J} &=
  \left[ {\begin{array}{cc}
   \mathbf{K}_{X_R^m} & \mathbf{K}_{X_R^m, X_I^m} \\
   \mathbf{K}_{X_I^m, X_R^m} & \mathbf{K}_{X_I^m} \\
  \end{array} } \right], \\
  \mathbf{\Sigma_S} &=
  \left[ {\begin{array}{cc}
   \mathbf{K}_{X_R^m} & \overline{\mathbf{K}}_{X_R^m, X_I^m} \\
   \overline{\mathbf{K}}_{X_I^m, X_R^m} & \mathbf{K}_{X_I^m} \\
  \end{array} } \right], \\
  \mathbf{K}_{X_R^m, X_I^m} &=
   \left[ \begin{array}{cc}
        \mathbf{K}_{Z_R^m, Z_I^m} & \mathbf{K}_{Z_R^m, \Lambda_I^m} \\
        \mathbf{K}_{\Lambda_R^m, Z_I^m} & \mathbf{K}_{\Lambda_R^m, \Lambda_I^m} \\
    \end{array} \right], \\
  \overline{\mathbf{K}}_{X_R^m, X_I^m} &= 
    \left[ \begin{array}{cc}
        \mathbf{0}_{kN \times kN} & \mathbf{K}_{Z_R^m, \Lambda_I^m} \\
        \mathbf{K}_{\Lambda_R^m, Z_I^m} & \mathbf{K}_{\Lambda_R^m, \Lambda_I^m} \\
    \end{array} \right].
\end{align}
We note that $\mathbf{K}_{Z_R^m, Z_I^m} = \mathbf{0}_{dN \times dN}$ in the separate-then-joint processing approach, since $f(\cdot)$ and $g(\cdot)$ do not consider the interactions between $Z_R^m$ and $Z_I^m$. This is characterized in Section \ref{sec:complementarity_principle} of the Appendix. To measure the magnitude of the data interactions across both approaches, we consider the $L_{p,q}$ norm, with $p,q \geq 1$, of $\mathbf{\Sigma_J}$ and $\mathbf{\Sigma_S}$. As $\mathbf{\Sigma_J}$ includes the aforementioned cross-covariance matrix of $Z_R^m$ and $Z_I^m$, it follows that $\|\mathbf{\Sigma_J}\|_{p,q} \geq \|\mathbf{\Sigma_S}\|_{p,q}$, as shown in Corollary \ref{corr:complementarity}.
\begin{corollary} \label{corr:complementarity}
Let $\mathbf{\Sigma_J}$ be the matrix of covariances of $X_R^m$, $X_I^m$ from joint-only processing, and let $\mathbf{\Sigma_S}$ be the matrix of covariances of $X_R^m$, $X_I^m$ from separate-then-joint processing. It follows that:
\begin{align}
    &\|\mathbf{\Sigma_J}\|_{p,q} \geq \|\mathbf{\Sigma_S}\|_{p,q}, \quad \text{where:} \\
    &Z_R^m \ind Z_I^m \implies \|\mathbf{\Sigma_J}\|_{p,q} = \|\mathbf{\Sigma_S}\|_{p,q}.
\end{align}
\end{corollary}
This greater norm indicates reduced interpretability, as the presence of cross-covariance terms implies that joint-only processing must not only handle $Z_R^m$ and $Z_I^m$ individually, but also their interactions, which can complicate the representation process. As such, $\mathbf{\Sigma_S}$ has a smaller \(L_{p,q}\) norm and associated representational complexity, enabling the Steinmetz network to separately extract $Z_R^m$ and $Z_I^m$ without the added burden of accounting for interactions. This Steinmetz architecture can also be leveraged to obtain a smaller upper bound on the generalization gap, as per Section \ref{sec:consistency_constraint}.

\section{Consistency Constraint} \label{sec:consistency_constraint}
Suppose $\psi_S$, $f_S$, and $g_S$ are stochastic transformations, where $\theta_{\psi_S} = (\theta_{f_S}, \theta_{g_S})$ is a random variable denoting the parameters of $\psi_S$, with the variables, $\theta_{f_S}$ and $\theta_{g_S}$, parameterizing $f_S$ and $g_S$, respectively. Per \citep{Federici2020Learning, fischer2020, lee2021compressive}, obtaining an optimal latent representation, $Z^m$, can be formulated as minimizing the mutual information between $X^m$ and $Z^m$, conditioned on $Y^m$. However, as explored by \citep{kolahi2020}, this framework does not hold when the encoder, $\psi_s$, is learned with the training dataset, $s$. To avoid this counterexample, \citep{kawaguchi23a} proposed an additional term that captures the amount of information in $S$ that is used to train the encoder, $\psi_S$. As such, in the context of our Steinmetz neural network architecture, obtaining the optimal $Z^m$ can be found by minimizing the expression in Eq. (\ref{eq:optimal_Z}), where $\theta_{\psi_S} \in \mathbb{R}^c, \theta_{f_S} \in \mathbb{R}^{c_1}, \theta_{g_S} \in \mathbb{R}^{c_2}$.
\begin{align} \label{eq:optimal_Z}
    \mathcal{J}(Z^m) &= I(X^m ; Z^m | Y^m) + I(S ; \theta_{\psi_S}) \\
    \notag &= I(X^m ; Z^m) - I(Y^m ; Z^m) + I(S ; \theta_{\psi_S})
\end{align}
Per Eq~(\ref{eq:optimal_Z}), the optimal latent representation, $Z^m$, best captures relevant information from $X^m$ about $Y^m$ while also considering the influence of $S$ on the encoder parameters, $\theta_{\psi_S}$. We now consider the upper bound on the generalization gap over the training dataset, $\Delta s$, which is an adapted version of the bound originally proposed in \citep{kawaguchi23a}.
\begin{theorem} \label{theorem:generalization_error}
For any $\delta > 0$ with probability at least $1 - \delta$ over the training dataset, $s$, we have that:
\begin{align} \label{eq:growth_rate}
    \notag&\Delta s = \Bigg[\mathbb{E}\big[\ell(\hat{Y}^m, Y^m)\big] -\frac{1}{M}\sum_{m \in \mathcal{V}} \ell(\hat{y}^m, y^m) \Bigg] \leq K(Z^m), \\
    &\text{where: } K(Z^m) = \frac{K_1(\alpha)}{\sqrt{M}} +
    K_2 \sqrt{\frac{\mathcal{J}(Z^m)\log(2) + K_3}{M}}.
\end{align}
The complete formulas for $K_1(\alpha)$, $K_2$, and $K_3$ can be found in Section \ref{sec:consistency_derivation} of the Appendix. We note that $\ell : \mathcal{Y} \times \mathcal{Y} \rightarrow \mathbb{R}_{\geq 0}$ is a bounded per-sample loss function.
\end{theorem}
The upper bound from Eq.~(\ref{eq:growth_rate}) captures the tradeoff between how well the latent space encapsulates information about the labels, and how much the encoder overfits the training distribution, wherein smaller values of $[I(X^m ; Z^m | Y^m) + I(S;\theta_{\psi_S})]$ yield a smaller upper bound on the generalization gap. Accordingly, we pose the following inquiry: \textit{is it possible to leverage the Steinmetz neural network architecture to obtain a smaller upper bound on the generalization gap}? To this end, we establish the existence of a lower bound, $\mathcal{D}(Z^m)$, on $[I(X^m ; Z^m | Y^m) + I(S;\theta_{\psi_S})]$ that is achievable using a constraint on the latent space of the Steinmetz neural network. We formalize this in Corollary \ref{corr:bound}.
\begin{corollary} \label{corr:bound}
Suppose $\mathcal{F}^m$ denotes the set of all constraints on $Z^m$. We have that $\forall m \in \mathcal{V}$:
\begin{gather} \label{eq:conjecture_bound}
    \mathcal{D}(Z^m) \leq \mathcal{J}(Z^m), \ \forall Z^m \in \mathbb{C}^{lN} \\
    \exists f \in \mathcal{F}^m \, : \, \forall Z^m \in \mathcal{E}, \ \mathcal{D}(Z^m) = \mathcal{J}(Z^m).
\end{gather}
Where $\mathcal{E} = \{Z^m \in \mathbb{C}^{lN} | f \}$ denotes the set of all $Z^m$ satisfying the constraint $f \in \mathcal{F}^m$.
\end{corollary}
Achieving a smaller upper bound on the generalization gap is indicative of a network's potential for improved accuracy in making predictions on unseen data. This relationship is deeply rooted in the notions of statistical learning theory, and more formally, in Structural Risk Minimization (SRM) and VC theory \citep{vapnik1971uniform, vapnik1999overview}. Consequently, should there exist a consistency constraint on the latent space yielding a smaller upper bound on $\Delta s$, we would expect it to improve the Steinmetz neural network's capacity to generalize \citep{vapnik2013nature}.

We have proven Corollary \ref{corr:bound} in Section \ref{sec:consistency_derivation} of the Appendix, and present the lower bound, $\mathcal{D}(Z^m)$, in Theorem \ref{theorem:overall_lower_bound}, wherein there exists a consistency constraint ensuring the achievability of $\mathcal{D}(Z^m)$.
\begin{theorem} \label{theorem:overall_lower_bound}
Consider $X^m = X_R^m + iX_I^m \in \mathbb{C}^{dN}$, $Y^m \in \mathbb{C}^{k}$, and $Z^m = Z_R^m + iZ_I^m \in \mathbb{C}^{lN}$, with $\theta_{\psi_S} \in \mathbb{R}^c, \theta_{g_S} \in \mathbb{R}^{c_2}$. It follows that $\mathcal{D}(Z^m) \leq \mathcal{J}(Z^m)$, where:
\begin{align}
    \notag &\mathcal{D}(Z^m) = H(Z_R^m) - I(Y^m ; Z^m) - M \big[H(X^m | \theta_{\psi_S}) \\
    & \quad \quad - H(Y^m | X_R^m + iX_I^m, \theta_{g_S}) - H(Y^m) \\
    \notag &\quad \quad - H(X_R^m | Y^m) - H(iX_I^m | X_R^m, iZ_I^m, Y^m) \big].
\end{align}
With equality if the following condition holds $\forall m \in \mathcal{V}$:
\begin{align}
\begin{gathered}
    \forall Z_I^m \in \mathbb{R}^{lN}, \ \exists! Z_R^m \in \mathbb{R}^{lN} : \\ Z_I^m = \phi(Z_R^m) \implies
    \mathcal{J}(Z^m) = \mathcal{D}(Z^m).
\end{gathered}
\end{align}
\end{theorem}
Theorem \ref{theorem:overall_lower_bound} informs us that $\mathcal{D}(Z^m)$ is achievable when we enforce $Z_I^m = \phi(Z_R^m)$, where $\phi(\cdot)$ is a deterministic, bijective function. We note that as the Steinmetz neural network is trained to minimize the average loss on the training dataset (through empirical risk minimization), we expect $Z^m$ to become more informative about the labels, whereby $I(Y^m ; Z^m)$ increases. We now further extend this result to Theorem \ref{theorem:generalization_error}, via which we obtain a smaller upper bound on the generalization gap.
\begin{theorem} \label{theorem:generalization_error_smaller}
For any $\delta > 0$ with probability at least $1 - \delta$ over the training dataset, $s$, we have that:
\begin{align} \label{eq:growth_rate_new}
    \notag &\Delta s \leq G(Z^m) \leq K(Z^m), \\
    &\text{where: } G(Z^m) = \frac{K_1(\alpha)}{\sqrt{M}} + K_2 \sqrt{\frac{\mathcal{D}(Z^m)\log(2) + K_3}{M}}.
\end{align}
With $G(Z^m) = K(Z^m)$ if the following condition holds:
\begin{align}
\begin{gathered}
    \forall Z_I^m \in \mathbb{R}^{lN}, \ \exists! Z_R^m \in \mathbb{R}^{lN} : \\
    Z_I^m = \phi(Z_R^m) \implies G(Z^m) = K(Z^m).
\end{gathered}
\end{align}
The complete formulas for $K_1(\alpha)$, $K_2$, and $K_3$ can be found in Section \ref{sec:consistency_derivation} of the Appendix. 
\end{theorem}

\section{Analytic Neural Network} \label{sec:analytic_neural_network}
As detailed in Section \ref{sec:consistency_constraint}, by introducing a constraint within the latent representation, $Z^m$, such that $Z_R^m$ and $Z_I^m$, are related by a deterministic, bijective function, $\phi(\cdot)$, we can leverage improved control over the generalization gap, $\Delta s$. A natural question that follows is: \textit{which $\phi(\cdot)$ should be chosen to improve predictive performance}? To address this, we consider a configuration that focuses on the properties and predictive advantages of orthogonal latent representations.

In feature engineering literature, selecting features that are orthogonal to others is a common strategy to minimize redundancy and improve network performance \citep{chaudhry2020continual}. For our application, with $Z_R^m$ and $Z_I^m$ as the latent features, we aim to use a function, $\phi(\cdot)$, which ensures these features are as orthogonal as possible. This objective aligns with the above principle of using non-redundant features. We revisit the analytic signal construction from Section \ref{sec:analytic_signals}, wherein the real and imaginary parts are related by the DHT, and are orthogonal to each other. Accordingly, for $\phi(\cdot) = \mathcal{H}\{ \cdot \}$, where $Z^m = Z_R^m + iZ_I^m$, with $Z_I^m = \mathcal{H}\{ Z_R^m \}$, it follows that $Z_R^m$ and $Z_I^m$ are orthogonal (see Section \ref{sec:hilbert_orthogonality} of the Appendix). We formalize this in Corollary \ref{corr:hilbert_constraint}.
\begin{corollary} \label{corr:hilbert_constraint}
Consider $Z^m = Z_R^m + iZ_I^m \in \mathbb{C}^{lN}$, $\langle \cdot, \cdot \rangle : \mathcal{Z} \times \mathcal{Z} \rightarrow \mathbb{R}_{\geq 0}$. We have that $\forall m \in \mathcal{V}$:
\begin{equation}
    Z_I^m = \mathcal{H}\{ Z_R^m \} \implies \langle Z_R^m, Z_I^m \rangle = 0.
\end{equation}
\end{corollary}
We term this Steinmetz neural network, where $Z_I^m \rightarrow \mathcal{H}\{Z_R^m\}$ during training as the \textit{Analytic Neural Network}. In leveraging $\phi(\cdot) = \mathcal{H}(\cdot)$ as our consistency constraint, we provide a framework that encourages orthogonality between $Z_R^m$ and $Z_I^m$, aiming to improve the Steinmetz neural network's predictive capabilities. We outline the practical implementation of this consistency constraint through the \textit{Hilbert Consistency Penalty}.

Consider sets $\{x_R^m, \, m \in \mathcal{V}\}$ and $\{x_I^m, \, m \in \mathcal{V}\}$ from the training dataset, with $x_R^m, x_I^m \in \mathbb{R}^{dN}$. It follows that $z_R^m = g(x_R^m)$ and $z_I^m = f(x_I^m)$, wherein $\smash{z_R^m, z_I^m \in \mathbb{R}^{lN}}$. To implement the Hilbert consistency penalty, we make use of the discrete Fourier transform (DFT), leveraging its properties in relation to phase shifts --- we consider $F_R^m = \mathcal{F}\{ z_R^m \} \in \mathbb{C}^{lN}$ as the DFT of $z_R^m$ where $b$ denotes the frequency index. Eq.~(\ref{eq:phase_shift}) summarizes the frequency domain implementation of this phase shift.
\begin{align} \label{eq:phase_shift}
    &H_R^m[b] = 
    \begin{cases} 
    F_R^m[b] \cdot (-i) & \ \text{for} \ \ 0 < b < \frac{lN}{2} \\
    F_R^m[b] & \ \text{for} \ \ b = 0, \ \frac{lN}{2} \\
    F_R^m[b] \cdot (i) & \ \text{for} \ \ \frac{lN}{2} < b < lN
    \end{cases}, \\
    &\text{where: } F_R^m[b] = \sum_{n=0}^{lN-1} z_R^m[n] e^{-\frac{i2\pi bn}{lN}}.
\end{align}
Above, $H_R^m \in \mathbb{C}^{lN}$ denotes the frequency components of $\mathcal{H}\{z_R^m\}$. Applying the inverse FFT to $H_R^m$ yields the discrete Hilbert transform of $z_R^m$, $\mathcal{H}\{ z_R^m\}$, in the time domain, as detailed in Eq.~(\ref{eq:hilbert_time_domain}).
\begin{align} \label{eq:hilbert_time_domain}
\begin{split}
    \mathcal{H}\{z_R^m\}[n] &= \mathcal{F}^{-1}\{H_R^m \}[n] \\
    &= \frac{1}{lN} \sum_{b=0}^{lN-1} H_R^m[b] e^{\frac{i2\pi bn}{lN}}.
\end{split}
\end{align}
We implement the Hilbert consistency penalty by penalizing the average error between $\mathcal{H}\{z_R^m\}$ and $z_I^m$, which we denote as $\mathcal{L}_{\mathcal{H}}$, where $\ell_{\mathcal{H}}$ is the relevant error metric. This penalty summarized in Definition \ref{def:hilbert_penalty}.
\begin{definition} \label{def:hilbert_penalty}
Consider $z_R^m, z_I^m \in \mathbb{R}^{lN}$, and suppose $H_R^m = \mathcal{F}\{\mathcal{H}\{z_R^m\} \} \in \mathbb{C}^{lN}$. We have that:
\begin{align}
    &\mathcal{L}_{\mathcal{H}} = \frac{1}{M} \sum_{m=0}^{M-1} \ell_{\mathcal{H}} \left( \mathcal{H}\{z_R^m\}, z_I^m \right), \\
    &\text{where: } \mathcal{L}_{\mathcal{H}} = 0 \iff z_I^m = \mathcal{H}\{z_R^m\}, \, \forall m \in \mathcal{V}.
\end{align}
We note that $\ell_{\mathcal{H}} : \mathcal{Z} \times \mathcal{Z} \rightarrow \mathbb{R}_{\geq 0}$ is a bounded per-sample loss function.
\end{definition}
The cumulative loss function used to train the analytic neural network is derived as the weighted sum of the average loss on the training dataset and the Hilbert consistency penalty, where $\beta$ is the tradeoff parameter. During training, we jointly minimize the error on the training dataset, and encourage the network to form analytic signal representations in the latent space. The overall loss, $\mathcal{L}$, is given by:
\begin{align}
    &\mathcal{L} = \frac{1}{M} \sum_{m=0}^{M-1} \ell \left( \hat{y}^m, y^m \right) + \beta \mathcal{L}_{\mathcal{H}}.
\end{align}
Where the value of $\beta$ can be fine-tuned to optimize the predictive accuracy on the training dataset, $s$.

\begin{table*}[t!]
\caption{Test performance comparison on CV-MNIST $(M = 500)$, CV-CIFAR-10 $(M = 50{,}000)$, CV-CIFAR-100 $(M = 50{,}000)$, CV-FSDD $(M = 2{,}700)$, and RASPNet $(M = 20{,}000)$.}
\label{tab:classsification_acc}
\begin{center}
\begin{tabular}{c|c|c|c|c}
\hline 
& \textbf{\texttt{RVNN}} & \textbf{\texttt{CVNN}} & \textbf{\texttt{Steinmetz}} & \textbf{\texttt{Analytic}}\\
\textbf{Dataset} & \textbf{Test Perf.} & \textbf{Test Perf.} & \textbf{Test Perf.} & \textbf{Test Perf.} \\
\hline
CV-MNIST & $73.180 \scriptstyle \pm 0.407 \ (\%) $ & $71.716 \scriptstyle \pm 1.957 \ (\%)$ & $74.680 \scriptstyle \pm 0.722 \ (\%)$ & $75.580 \scriptstyle \pm 0.970 \ (\%)$ \\
CV-CIFAR-10 &  $42.734 \scriptstyle \pm 0.397 \ (\%)$ & $40.370 \scriptstyle \pm 0.417 \ (\%)$ & $44.922 \scriptstyle \pm 0.474 \ (\%)$ & $45.180 \scriptstyle \pm 0.239 \ (\%)$ \\
CV-CIFAR-100 &  $13.444 \scriptstyle \pm 0.197 \ \text{(\%)}$ & $12.412 \scriptstyle \pm 0.319 \ \text{(\%)}$ & $15.110 \scriptstyle \pm 0.137 \ \text{(\%)}$ & $15.380 \scriptstyle \pm 0.256 \ \text{(\%)}$ \\
CV-FSDD &  $20.067 \scriptstyle \pm 1.036 \ \text{(\%)}$ & $20.720 \scriptstyle \pm 1.006 \ \text{(\%)}$ & $24.600 \scriptstyle \pm 1.259 \ \text{(\%)}$ & $25.020 \scriptstyle \pm 0.749 \ \text{(\%)}$ \\
RASPNet &  $34867 \scriptstyle \pm 347.8 \ \text{(MSE)}$ & $36928 \scriptstyle \pm 197.9 \ \text{(MSE)}$ & $30360 \scriptstyle \pm 70.12 \ \text{(MSE)}$ & $29880 \scriptstyle \pm 169.9 \ \text{(MSE)}$ \\
\hline
\textbf{Dataset} & \textbf{Parameters} & \textbf{Parameters} & \textbf{Parameters} & \textbf{Parameters} \\
\hline 
CV-MNIST & $52{,}970$ & $55{,}764$ & $53{,}002$ & $53{,}002$ \\
CV-CIFAR-10 & $199{,}402$ & $202{,}196$ & $199{,}434$ & $199{,}434$ \\
CV-CIFAR-100 & $205{,}252$ & $213{,}896$ & $205{,}284$ & $205{,}284$ \\
CV-FSDD & $1{,}622{,}310$ & $1{,}644{,}620$ & $1{,}622{,}410$ & $1{,}622{,}410$ \\
RASPNet & $223{,}682$ & $232{,}324$ & $223{,}746$ & $223{,}746$ \\
\hline
\end{tabular}
\end{center}
\end{table*}
\begin{figure*}[t!]
    \centering
    \begin{subfigure}{0.32\textwidth}
        \includegraphics[width=\textwidth]{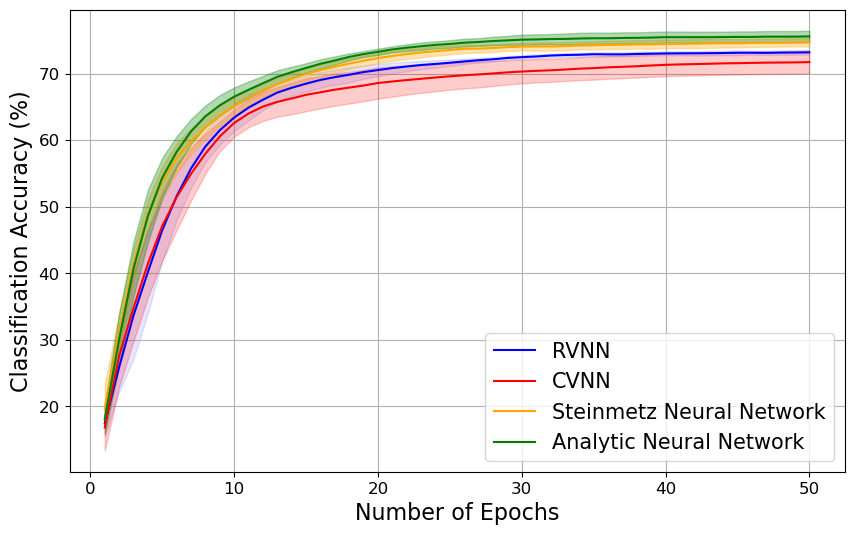}
        \caption{Complex-Valued MNIST}
    \end{subfigure}
    \begin{subfigure}{0.32\textwidth}
        \includegraphics[width=\textwidth]{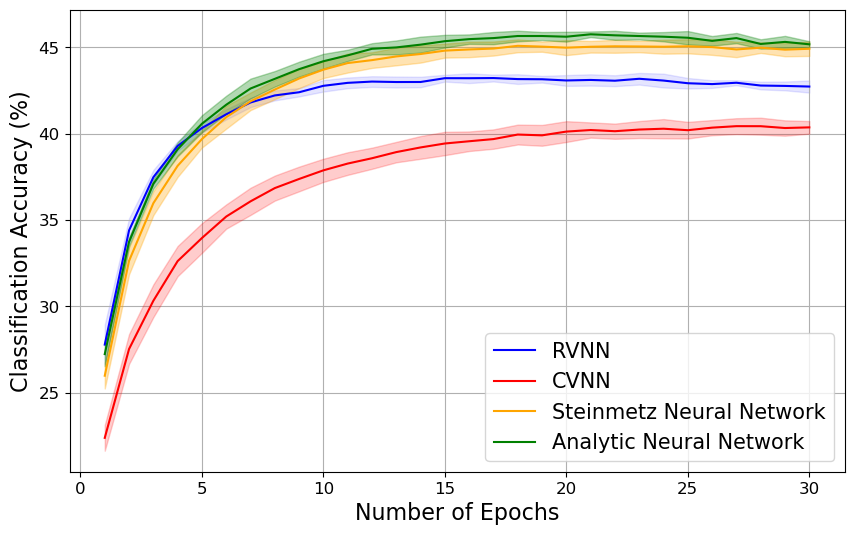}
        \caption{Complex-Valued CIFAR-10}
    \end{subfigure}
    \begin{subfigure}{0.32\textwidth}
        \includegraphics[width=\textwidth]{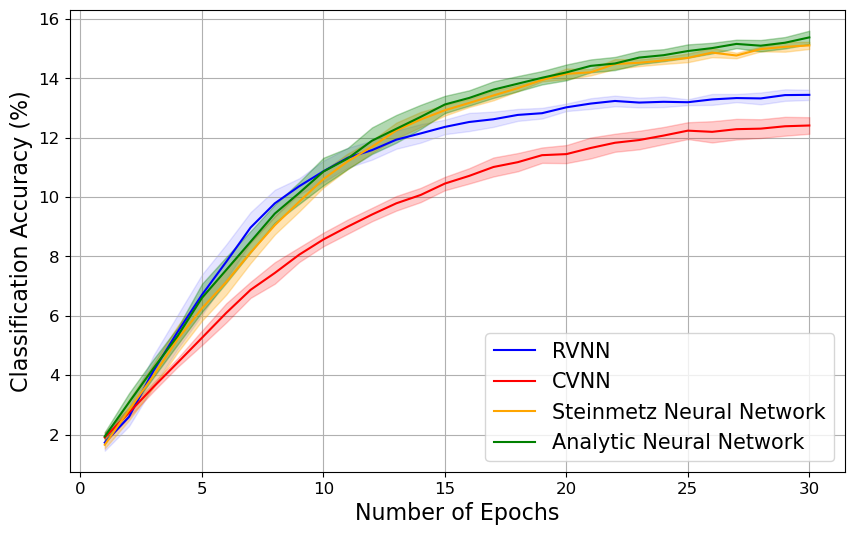}
        \caption{Complex-Valued CIFAR-100}
    \end{subfigure}
    \\
    \begin{subfigure}{0.32\textwidth}
        \includegraphics[width=\textwidth]{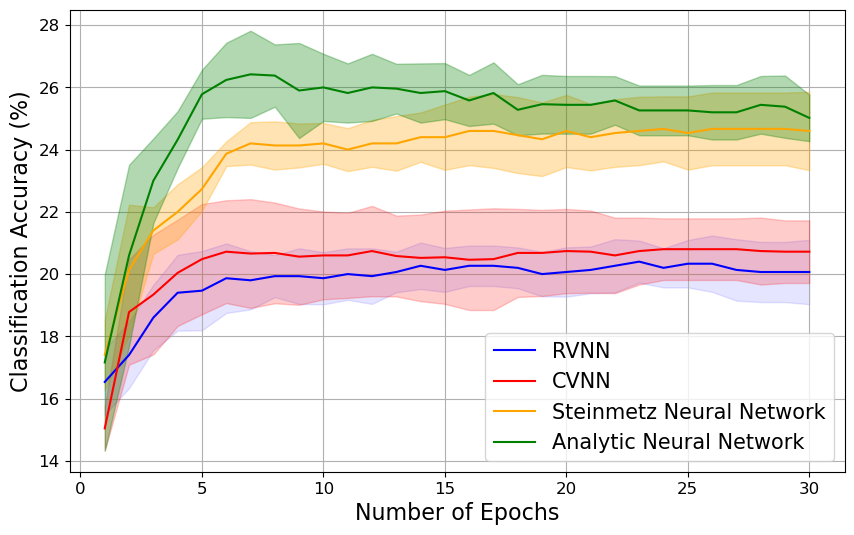}
        \caption{Complex-Valued FSDD}
    \end{subfigure}
    \begin{subfigure}{0.33\textwidth}
        \includegraphics[width=\textwidth]{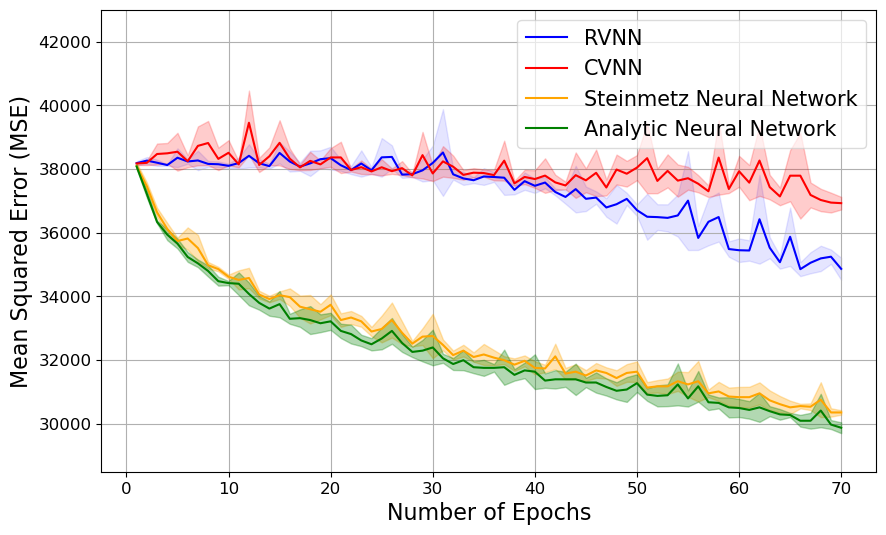}
        \caption{RASPNet Scenario $i = 29$}
    \end{subfigure}
    \caption{Test performance comparison on CV-MNIST $(M = 500)$, CV-CIFAR-10 $(M = 50{,}000)$, CV-CIFAR-100 $(M = 50{,}000)$, CV-FSDD $(M = 2{,}700)$, and RASPNet $(M = 20{,}000)$ through CVNN, RVNN, Steinmetz neural network, and analytic neural network. The x-axis indicates the training epochs, while the y-axis indicates the test performance (classification accuracy and mean squared error).}
    \label{fig:epoch_performance}
\end{figure*}

\section{Empirical Results} \label{sec:empirical_results}
We present empirical results on benchmark datasets for complex-valued multi-class classification and regression, and on a synthetic signal processing example for complex-valued regression. Per \citep{trabelsi2017deep} and \citep{scardapane2020complex}, we present classification results on complex-valued MNIST~\citep{deng2012mnist}, CIFAR-10~\citep{krizhevsky2009learning}, CIFAR-100~\citep{krizhevsky2009learning}, and FSDD~\citep{jacksonFSDD}. We also present a regression result on the RASPNet benchmark dataset \citep{venkatasubramanian2024raspnet}, which comprises intrinsically complex-valued radar returns. The CVNNs in this analysis were constructed through the Complex Pytorch library \citep{maxime2021disorder}, which implements the layers proposed in \citep{trabelsi2017deep}.

\begin{figure*}[h!]
    \centering
    \begin{subfigure}{0.48\textwidth}
        \includegraphics[width=\textwidth]{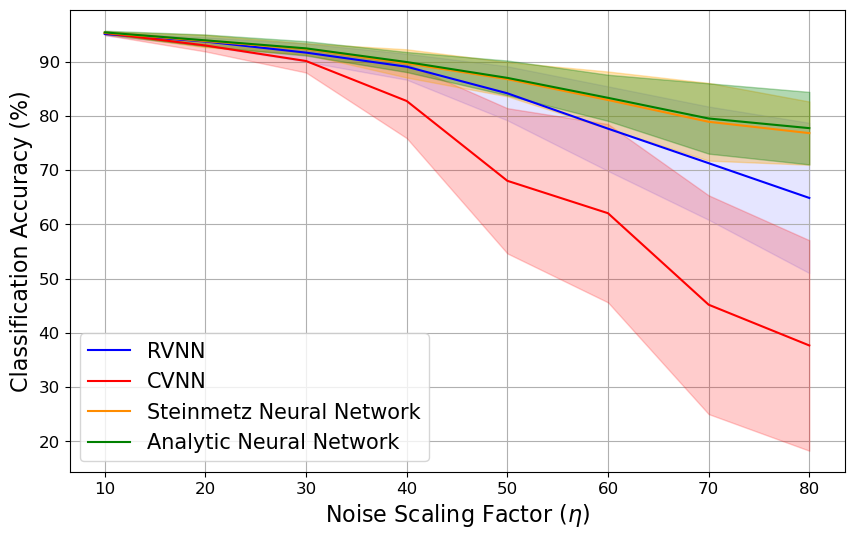}
        \caption{Noisy Complex-Valued MNIST}
    \end{subfigure}
    \begin{subfigure}{0.48\textwidth}
        \includegraphics[width=\textwidth]{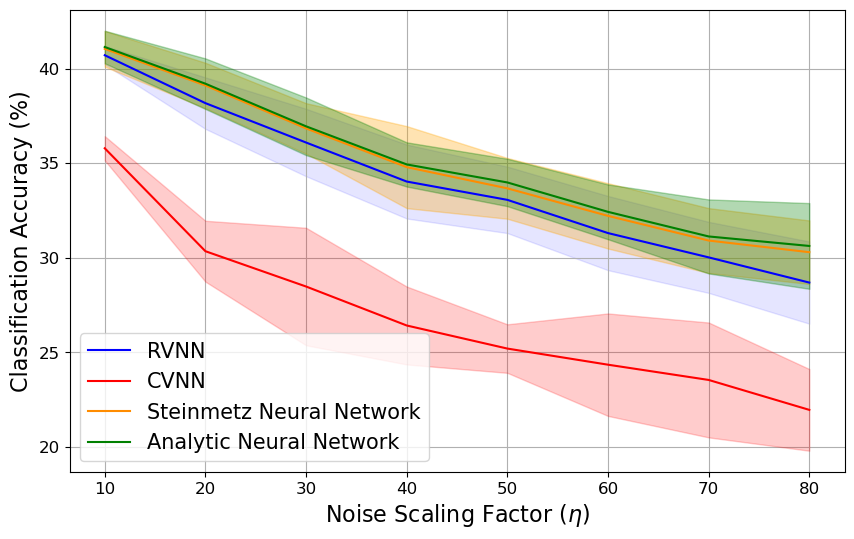}
        \caption{Noisy Complex-Valued CIFAR-10}
    \end{subfigure}
    \caption{Noise robustness test performance on CV-MNIST ($M = 60{,}000$) and CV-CIFAR-10 ($M = 50{,}000$) through CVNN, RVNN, Steinmetz neural network, and analytic neural network. The x-axis is the scaling factor, $\eta$, for the additive complex normal noise, while the y-axis indicates the classification accuracy.}
    \label{fig:noise_robustness}
\end{figure*}

\begin{table*}[t!]
\caption{Test MSE (magnitude and phase) for channel identification task with $\rho = \sqrt{2}/2$.}
\label{tab:regression_mse}
\begin{center}
\begin{tabular}{c|c|c|c|c}
\hline 
& \textbf{\texttt{RVNN}} & \textbf{\texttt{CVNN}} & \textbf{\texttt{Steinmetz}} & \textbf{\texttt{Analytic}} \\
\hline
Magnitude Error &  $1.035 \scriptstyle \pm 0.011$ & $1.003 \scriptstyle \pm 0.006$ & $1.151 \scriptstyle \pm 0.005$ & $1.114 \scriptstyle \pm 0.078$ \\
\hline 
Phase Error & $4.270 \scriptstyle \pm 0.095$ & $4.451 \scriptstyle \pm 0.085$ & $3.808 \scriptstyle \pm 0.129$ & $3.768 \scriptstyle \pm 0.169$ \\
\hline
Parameters & $2{,}594$ & $4{,}738$ & $2{,}626$ & $2{,}626$ \\
\hline
\end{tabular}
\end{center}
\end{table*}

\subsection{Benchmark Datasets} \label{sec:benchmark_datasets}
The first experiment we consider is an assessment of our methods on Complex-Valued MNIST (CV-MNIST), Complex-Valued CIFAR-10 (CV-CIFAR-10), Complex-Valued CIFAR-100 (CV-CIFAR-100), and the audio-based Complex-Valued Free Spoken Digit Dataset (CV-FSDD), for multi-class classification, and on RASPNet for regression, evaluating the efficacy of the proposed Steinmetz and analytic neural networks when there are no ablations introduced within the data. We obtain CV-MNIST by taking a $dN = 784$-point DFT of each of the $M = 500$ training images (a small subset of the first $500$ images in MNIST). The test set is formed by taking a $784$-point DFT of each of the $10{,}000$ test images. We obtain CV-CIFAR-10 and CV-CIFAR-100 by taking a $dN = 3072$-point DFT of each of the $M = 50{,}000$ training images. The test set is constructed by taking a $3072$-point DFT of each of the $10{,}000$ test images. We obtain CV-FSDD by taking a $dN = 8000$-point DFT of each of the $M = 2{,}700$ training audio signals. The test set is constructed by taking a $8000$-point DFT of each of the $300$ test audio signals. For CV-MNIST, CV-CIFAR-10, and CV-FSDD we have $k = 10$, and for CV-CIFAR-100, we have $k = 100$. RASPNet consists of $M = 20{,}000$ training and $5{,}000$ test radar returns of size $dN = 1680$, with $k = 2$ ($2$D target localization).

We train a RVNN, CVNN, Steinmetz neural network, and analytic neural network using Cross Entropy Loss for $lN = 64$ to classify the images in CV-MNIST, CV-CIFAR-10, and CV-CIFAR-100, $lN = 200$ to classify the audio signals in CV-FSDD, and using MSE Loss for $lN = 128$ to localize targets from the radar returns in RASPNet. These neural network architectures and hyperparameter choices are described in Section \ref{sec:network_architectures} of the Appendix, wherein we leverage the Adam optimizer \citep{kingma2014adam} to train each architecture using a fixed learning rate. The empirical results pertaining to this first experiment are depicted in Table \ref{tab:classsification_acc} and Figure \ref{fig:epoch_performance}. On CV-MNIST, CV-CIFAR-10, CV-CIFAR-100, CV-FSDD, and RASPNet, we observe that the Steinmetz and analytic neural networks achieve improved generalization over the classical RVNN and the CVNN. The analytic neural network achieves the highest classification accuracy and the lowest MSE.

The second experiment is an examination of the impact of additive complex normal noise on the performance of our proposed neural networks on CV-MNIST and CV-CIFAR-10. We add standard complex normal noise scaled by a factor, $\eta$, to each example $x^m \in s$, where $M = 60{,}000$ for CV-MNIST and $M = 50{,}000$ for CV-CIFAR-10. The new training dataset, $s'$, is given by:
\begin{align}
    &s' = \{(x^{m'},y^m), m \in \mathcal{V} \}, \\
    &\text{where: } x^{m'} = x^m + \eta \times \mathcal{CN}(\mathbf{0}, \mathbf{I}_{dN}).
\end{align}
This experimental setup allows us to gauge how the signal-to-noise ratio (SNR) influences the efficacy of our Steinmetz and analytic neural networks. The empirical results for this experiment are given in Figure \ref{fig:noise_robustness}. We see that across both datasets, the Steinmetz and analytic neural networks are far more resilient to additive noise.

\subsection{Channel Identification}
We evaluate our proposed Steinmetz and analytic networks on the benchmark channel identification task from \citep{scardapane2020complex, bouboulis2015complex}. Let $\smash{X^m = \sqrt{1 - \rho^2} \bar{X}\vphantom{X}^m + i \rho \tilde{X}\vphantom{X}^m}$ denote the input to the channel, wherein $\smash{\bar{X}\vphantom{X}^m}$ and $\smash{\tilde{X}\vphantom{X}^m}$ are Gaussian random variables, and $\rho$ determines the circularity of the signal. Here, $dN = 5$ denotes the length of the input sequence, an embedding of the channel inputs over $dN$ time steps. The channel output, $Y^m \in \mathbb{C}^k$, is formed using a linear filter, a memoryless nonlinearity, and by adding white Gaussian noise to achieve an SNR of $5$ dB, wherein $X^m$ and $Y^m$ are equivalent to $s_n$ and $r_n$ from \citep{scardapane2020complex}. We consider $M = 1000$ training examples and $1000$ test examples. Each of the real-valued architectures output a $2k = 2$D vector, $\smash{[[\hat{Y}_R^m]^T, [\hat{Y}_I^m]^T]^T}$, where $\smash{\hat{Y}_R^m, \hat{Y}_I^m \in \mathbb{R}}$, and the CVNN outputs a $k = 1$-dimensional scalar, $\smash{\hat{Y}^m = \hat{Y}_R^m + i\hat{Y}_I^m}$, which we reshape to form $\smash{[[\hat{Y}_R^m]^T, [\hat{Y}_I^m]^T]^T}$. We train our Steinmetz and analytic neural networks, RVNN, and CVNN using MSE Loss, minimizing the distance between $\smash{[[\hat{Y}_R^m]^T, [\hat{Y}_I^m]^T]^T}$ and $\smash{[[Y_R^m]^T, [Y_I^m]^T]^T}$. We select $lN = 64$, and use the Adam optimizer to train each architecture using a fixed learning rate. We compute and report the MSE between the predicted and true magnitudes and phases on the test dataset in Table \ref{tab:regression_mse}. From Table \ref{tab:regression_mse}, we see while the magnitude prediction error is comparable between the CVNN, RVNN, and the Steinmetz and analytic neural networks, the latter pair observes a much lower phase prediction error. 

\section{Conclusion} \label{sec:conclusion}
In this work, we introduced Steinmetz neural networks, a new approach to processing complex-valued data using DNNs with parallel real-valued subnetworks. We provided its mathematical framework and outlined a consistency constraint to lower its generalization gap upper bound, and we presented the analytic neural network, which incorporates the consistency penalty, for practical implementation. We evaluated these networks on regression and classification benchmarks, showing improvements over existing RVNNs and CVNNs. Future work includes investigating more effective training techniques for Steinmetz neural networks and theoretical performance guarantees.

\subsubsection*{Acknowledgements}
This work was supported in part by the U.S. Air Force Office of Scientific Research (AFOSR) under award FA9550-21-1-0235. Any opinions, findings and conclusions expressed in this material are those of the authors and do not necessarily reflect the views of the U.S. Department of Defense.

\bibliographystyle{plainnat}
\bibliography{references}

\begin{thebibliography}{37}
\providecommand{\natexlab}[1]{#1}
\providecommand{\url}[1]{\texttt{#1}}
\expandafter\ifx\csname urlstyle\endcsname\relax
  \providecommand{\doi}[1]{doi: #1}\else
  \providecommand{\doi}{doi: \begingroup \urlstyle{rm}\Url}\fi

\bibitem[Barrachina et~al.(2021)Barrachina, Ren, Morisseau, Vieillard, and Ovarlez]{barrachina2021cvnn}
J.~A. Barrachina, C.~Ren, C.~Morisseau, G.~Vieillard, and J.-P. Ovarlez.
\newblock Complex-valued vs. real-valued neural networks for classification perspectives: An example on non-circular data.
\newblock In \emph{ICASSP 2021 - 2021 IEEE International Conference on Acoustics, Speech and Signal Processing (ICASSP)}, pages 2990--2994, 2021.
\newblock \doi{10.1109/ICASSP39728.2021.9413814}.

\bibitem[Bassey et~al.(2021)Bassey, Qian, and Li]{bassey2021survey}
Joshua Bassey, Lijun Qian, and Xianfang Li.
\newblock A survey of complex-valued neural networks, 2021.

\bibitem[Bouboulis et~al.(2015)Bouboulis, Theodoridis, Mavroforakis, and Evaggelatou-Dalla]{bouboulis2015complex}
Pantelis Bouboulis, Sergios Theodoridis, Charalampos Mavroforakis, and Leoni Evaggelatou-Dalla.
\newblock Complex support vector machines for regression and quaternary classification.
\newblock \emph{IEEE Transactions on Neural Networks and Learning Systems}, 26\penalty0 (6):\penalty0 1260--1274, 2015.
\newblock \doi{10.1109/TNNLS.2014.2336679}.

\bibitem[Chaudhry et~al.(2020)Chaudhry, Khan, Dokania, and Torr]{chaudhry2020continual}
Arslan Chaudhry, Naeemullah Khan, Puneet Dokania, and Philip Torr.
\newblock Continual learning in low-rank orthogonal subspaces.
\newblock \emph{Advances in Neural Information Processing Systems}, 33:\penalty0 9900--9911, 2020.

\bibitem[Cizek(1970)]{cizek1970hilbert}
V.~Cizek.
\newblock Discrete hilbert transform.
\newblock \emph{IEEE Transactions on Audio and Electroacoustics}, 18\penalty0 (4):\penalty0 340--343, 1970.
\newblock \doi{10.1109/TAU.1970.1162139}.

\bibitem[Deng(2012)]{deng2012mnist}
Li~Deng.
\newblock The mnist database of handwritten digit images for machine learning research.
\newblock \emph{IEEE Signal Processing Magazine}, 29\penalty0 (6):\penalty0 141--142, 2012.

\bibitem[Federici et~al.(2020)Federici, Dutta, Forré, Kushman, and Akata]{Federici2020Learning}
Marco Federici, Anjan Dutta, Patrick Forré, Nate Kushman, and Zeynep Akata.
\newblock Learning robust representations via multi-view information bottleneck.
\newblock In \emph{International Conference on Learning Representations}, 2020.

\bibitem[Fischer(2020)]{fischer2020}
Ian Fischer.
\newblock The conditional entropy bottleneck.
\newblock \emph{Entropy}, 22\penalty0 (9), 2020.
\newblock ISSN 1099-4300.
\newblock \doi{10.3390/e22090999}.

\bibitem[Gao et~al.(2019)Gao, Deng, Qin, Wang, and Li]{gao2019radar}
Jingkun Gao, Bin Deng, Yuliang Qin, Hongqiang Wang, and Xiang Li.
\newblock Enhanced radar imaging using a complex-valued convolutional neural network.
\newblock \emph{IEEE Geoscience and Remote Sensing Letters}, 16\penalty0 (1):\penalty0 35--39, 2019.
\newblock \doi{10.1109/LGRS.2018.2866567}.

\bibitem[Guberman(2016)]{guberman2016complex}
Nitzan Guberman.
\newblock On complex valued convolutional neural networks.
\newblock \emph{arXiv preprint arXiv:1602.09046}, 2016.

\bibitem[Guo et~al.(2021)Guo, Song, and Wu]{guo2021electromagnetic}
Liang Guo, Guanfeng Song, and Hongsheng Wu.
\newblock Complex-valued pix2pix—deep neural network for nonlinear electromagnetic inverse scattering.
\newblock \emph{Electronics}, 10\penalty0 (6), 2021.
\newblock ISSN 2079-9292.
\newblock \doi{10.3390/electronics10060752}.

\bibitem[Hafez-Kolahi et~al.(2020)Hafez-Kolahi, Kasaei, and Soleymani-Baghshah]{kolahi2020}
Hassan Hafez-Kolahi, Shohreh Kasaei, and Mahdiyeh Soleymani-Baghshah.
\newblock Sample complexity of classification with compressed input.
\newblock \emph{Neurocomputing}, 415:\penalty0 286--294, 2020.
\newblock ISSN 0925-2312.
\newblock \doi{https://doi.org/10.1016/j.neucom.2020.07.043}.

\bibitem[Hirose(2012)]{hirose2012complex}
Akira Hirose.
\newblock Complex-valued neural networks.
\newblock \emph{Springer Science \& Business Media}, 2012.

\bibitem[Jackson et~al.(2018)Jackson, Souza, Flaks, Pan, Nicolas, and Thite]{jacksonFSDD}
Zohar Jackson, César Souza, Jason Flaks, Yuxin Pan, Hereman Nicolas, and Adhish Thite.
\newblock Jakobovski/free-spoken-digit-dataset: v1.0.8 (version v1.0.8)., 2018.

\bibitem[Kawaguchi et~al.(2023)Kawaguchi, Deng, Ji, and Huang]{kawaguchi23a}
Kenji Kawaguchi, Zhun Deng, Xu~Ji, and Jiaoyang Huang.
\newblock How does information bottleneck help deep learning?
\newblock In \emph{Proceedings of the 40th International Conference on Machine Learning}, volume 202 of \emph{Proceedings of Machine Learning Research}, pages 16049--16096. PMLR, 23--29 Jul 2023.

\bibitem[Kingma and Ba(2014)]{kingma2014adam}
Diederik~P Kingma and Jimmy Ba.
\newblock Adam: A method for stochastic optimization.
\newblock \emph{arXiv preprint arXiv:1412.6980}, 2014.

\bibitem[Krizhevsky et~al.(2009)Krizhevsky, Hinton, et~al.]{krizhevsky2009learning}
Alex Krizhevsky, Geoffrey Hinton, et~al.
\newblock Learning multiple layers of features from tiny images.
\newblock 2009.

\bibitem[Lahat et~al.(2015)Lahat, Adalı, and Jutten]{lahat2015multimodal}
Dana Lahat, T{\"u}lay Adalı, and Christian Jutten.
\newblock Multimodal data fusion: An overview of methods, challenges, and prospects.
\newblock \emph{Proceedings of the IEEE}, 103\penalty0 (9):\penalty0 1449--1477, 2015.

\bibitem[Lee et~al.(2022)Lee, Hasegawa, and Gao]{lee2022survey}
ChiYan Lee, Hideyuki Hasegawa, and Shangce Gao.
\newblock Complex-valued neural networks: A comprehensive survey.
\newblock \emph{IEEE/CAA Journal of Automatica Sinica}, 9\penalty0 (8):\penalty0 1406--1426, 2022.
\newblock \doi{10.1109/JAS.2022.105743}.

\bibitem[Lee et~al.(2021)Lee, Arnab, Guadarrama, Canny, and Fischer]{lee2021compressive}
Kuang-Huei Lee, Anurag Arnab, Sergio Guadarrama, John Canny, and Ian Fischer.
\newblock Compressive visual representations.
\newblock In A.~Beygelzimer, Y.~Dauphin, P.~Liang, and J.~Wortman Vaughan, editors, \emph{Advances in Neural Information Processing Systems}, 2021.

\bibitem[Matth\`es et~al.(2021)Matth\`es, Bromberg, de~Rosny, and Popoff]{maxime2021disorder}
Maxime~W. Matth\`es, Yaron Bromberg, Julien de~Rosny, and S\'ebastien~M. Popoff.
\newblock Learning and avoiding disorder in multimode fibers.
\newblock \emph{Phys. Rev. X}, 11:\penalty0 021060, Jun 2021.
\newblock \doi{10.1103/PhysRevX.11.021060}.
\newblock URL \url{https://link.aps.org/doi/10.1103/PhysRevX.11.021060}.

\bibitem[Scardapane et~al.(2020)Scardapane, Van~Vaerenbergh, Hussain, and Uncini]{scardapane2020complex}
Simone Scardapane, Steven Van~Vaerenbergh, Amir Hussain, and Aurelio Uncini.
\newblock Complex-valued neural networks with nonparametric activation functions.
\newblock \emph{IEEE Transactions on Emerging Topics in Computational Intelligence}, 4\penalty0 (2):\penalty0 140--150, 2020.
\newblock \doi{10.1109/TETCI.2018.2872600}.

\bibitem[Shafran et~al.(2018)Shafran, Bagby, and Skerry-Ryan]{ShafranBS18}
Izhak Shafran, Tom Bagby, and R.~J. Skerry-Ryan.
\newblock Complex evolution recurrent neural networks (cernns).
\newblock In \emph{2018 IEEE International Conference on Acoustics, Speech and Signal Processing, ICASSP 2018, Calgary, AB, Canada, April 15-20, 2018}, pages 5854--5858. IEEE, 2018.
\newblock ISBN 978-1-5386-4658-8.
\newblock \doi{10.1109/ICASSP.2018.8462556}.

\bibitem[Smith(2023)]{smith2023complexvalued}
Josiah~W. Smith.
\newblock Complex-valued neural networks for data-driven signal processing and signal understanding, 2023.

\bibitem[Steinmetz(1893)]{Steinmetz1893}
Charles~Proteus Steinmetz.
\newblock \emph{Theory and Calculation of Alternating Current Phenomena}.
\newblock American Institute of Electrical Engineers, 1893.

\bibitem[Sun(2013)]{sun2013survey}
Shiliang Sun.
\newblock A survey of multi-view machine learning.
\newblock \emph{Neural Computing and Applications}, 23\penalty0 (7-8):\penalty0 2031--2038, 2013.

\bibitem[Tishby and Zaslavsky(2015)]{Tishby2015}
Naftali Tishby and Noga Zaslavsky.
\newblock Deep learning and the information bottleneck principle.
\newblock In \emph{IEEE Information Theory Workshop (ITW)}. IEEE, 2015.

\bibitem[Trabelsi et~al.(2018)Trabelsi, Bilaniuk, Zhang, Serdyuk, Subramanian, Santos, Mehri, Rostamzadeh, Bengio, and Pal]{trabelsi2017deep}
Chiheb Trabelsi, Olexa Bilaniuk, Ying Zhang, Dmitriy Serdyuk, Sandeep Subramanian, Jo{\~a}o~Felipe Santos, Soroush Mehri, Negar Rostamzadeh, Yoshua Bengio, and Christopher~J Pal.
\newblock Deep complex networks.
\newblock \emph{International Conference on Learning Representations}, 2018.

\bibitem[Vapnik and Chervonenkis(1971)]{vapnik1971uniform}
V.~N. Vapnik and A.~Y. Chervonenkis.
\newblock On the uniform convergence of relative frequencies of events to their probabilities.
\newblock \emph{Theory of Probability \& Its Applications}, 16\penalty0 (2):\penalty0 264--280, 1971.

\bibitem[Vapnik(2013)]{vapnik2013nature}
Vladimir Vapnik.
\newblock \emph{The nature of statistical learning theory}.
\newblock Springer science \& business media, 2013.

\bibitem[Vapnik(1999)]{vapnik1999overview}
Vladimir~N Vapnik.
\newblock An overview of statistical learning theory.
\newblock \emph{IEEE transactions on neural networks}, 10\penalty0 (5):\penalty0 988--999, 1999.

\bibitem[Venkatasubramanian et~al.(2024)Venkatasubramanian, Kang, Pezeshki, Rangaswamy, and Tarokh]{venkatasubramanian2024raspnet}
Shyam Venkatasubramanian, Bosung Kang, Ali Pezeshki, Muralidhar Rangaswamy, and Vahid Tarokh.
\newblock Raspnet: A benchmark dataset for radar adaptive signal processing applications.
\newblock \emph{arXiv preprint arXiv:2406.09638}, 2024.

\bibitem[Virtue et~al.(2017)Virtue, Yu, and Lustig]{Virtue2017BetterTR}
Patrick Virtue, Stella~X. Yu, and Michael Lustig.
\newblock Better than real: Complex-valued neural nets for mri fingerprinting.
\newblock \emph{2017 IEEE International Conference on Image Processing (ICIP)}, pages 3953--3957, 2017.

\bibitem[Wu et~al.(2023)Wu, Zhang, Jiang, and Zhou]{wu2023complex}
Jin-Hui Wu, Shao-Qun Zhang, Yuan Jiang, and Zhi-Hua Zhou.
\newblock Complex-valued neurons can learn more but slower than real-valued neurons via gradient descent.
\newblock pages 23714--23747, 2023.

\bibitem[Xu et~al.(2013)Xu, Tao, and Xu]{xu2013survey}
Chang Xu, Dacheng Tao, and Chao Xu.
\newblock A survey on multi-view learning.
\newblock \emph{arXiv preprint arXiv:1304.5634}, 2013.

\bibitem[Yan et~al.(2021)Yan, Hu, Mao, Ye, and Yu]{yan2021deep}
Xiaoqiang Yan, Shizhe Hu, Yiqiao Mao, Yangdong Ye, and Hui Yu.
\newblock Deep multi-view learning methods: A review.
\newblock \emph{Neurocomputing}, 448:\penalty0 106--129, 2021.
\newblock ISSN 0925-2312.
\newblock \doi{https://doi.org/10.1016/j.neucom.2021.03.090}.

\bibitem[Zhao et~al.(2017)Zhao, Xie, Xu, and Sun]{zhao2017multi}
Jing Zhao, Xing Xie, Xin Xu, and Shiliang Sun.
\newblock Multi-view learning overview: Recent progress and new challenges.
\newblock \emph{Information Fusion}, 38:\penalty0 43--54, 2017.

\end{thebibliography}

 \clearpage
\appendix
\onecolumn

\aistatstitle{Appendix}

\section{Consistency Constraint Derivation} \label{sec:consistency_derivation}
As outlined in Section \ref{sec:consistency_constraint}, our aim is to exploit the Steinmetz neural network architecture by deriving a consistency constraint that allows for improved control over the generalization gap, $\Delta s$. Recall Eq.~\ref{eq:growth_rate}, which provides an upper bound on $\Delta s$ in terms of the mutual information between $X^m$ and $Z^m$, between $Y^m$ and $Z^m$, and between $S$ and $\theta_{\psi_S}$. The complete derivation of this bound is provided in \citep{kawaguchi23a}.
\begin{lemma}
For any $\lambda > 0$, $\gamma > 0$ and $\delta > 0$ with probability at least $1 - \delta$ over $s$, we have that:
\begin{align}
    &\Delta s \leq K(Z^m) = \frac{K_1(\alpha)}{\sqrt{M}} + K_2 \sqrt{\frac{[I(X^m;Z^m) - I(Y^m;Z^m) + I(S;\theta_{\psi_S})]\log(2) + K_3}{M}}.
\end{align}
Where $\alpha = (I(\theta_{\psi_S} ; S) + K_4)\log(2) + \log(2)$, and:
\begin{align}
    &K_1(\alpha) = \frac{\max\limits_{m \in \mathcal{V}} \ell(\hat{y}^m, y^m) \sqrt{2\gamma|\mathcal{Y}^m|}}{M^{1/4}} \sqrt{\alpha + \log(2|\mathcal{Y}| / \delta)} + \gamma K_5, \\
    &K_2 = \max_{y^m \in \mathcal{Y}}  \sum_{k=1}^{M_{y^m}}\ell((\mathfrak{z}_k^{y^m})^m, y^m) \sqrt{2 | \mathcal{Y}|\mathbb{P}(Z^m = (\mathfrak{z}_{k}^{y^m})^m | Y = y^m)}, \\
    &K_3 = \bigg(\mathbb{E}_{y^m}[c_{y^m}(\theta_{\psi_s})] \sqrt{\Big[p \log(\sqrt{M / \gamma)}\Big]/2} + K_4 \bigg) \log(2).
\end{align}
Above, for $K_2$, $M_{y^m}$ denotes the size of the typical subset of the set of latent variables per $y^m \in \mathcal{Y}$, wherein the elements of the typical subset are given by: $\{(\mathfrak{z}_1^{y^m})^m,...,(\mathfrak{z}_{M_{y^m}}^{y^m})^m\}$. For $K_3$, $c_{y^m}(\theta_{\psi_s})$ denotes the sensitivity of $\theta_{\psi_s}$, $\theta_{\psi_s} \in \mathbb{R}^c$, and $K_4 = \frac{1}{\lambda} \log \big(\frac{1}{\delta e^{\lambda H(\theta_{\psi_S})}} \sum_{q \in \mathcal{M}}(\mathbb{P}(\theta_{\psi_S} = q))^{1 - \lambda} \big) + H(\theta_{\psi_S} | S)$. Additionally, for $K_1$, we have that $K_5 = \max_{(x^m, y^m)\in (\mathcal{X} \times \mathcal{Y})} \ell(h(\psi_s(x^m)), y^m)$.
\end{lemma}

\noindent For improved control over the generalization gap, we form a smaller upper bound on $\Delta s$ by deriving a lower bound, $\mathcal{D}(Z^m)$, on $[I(X^m;Z^m) - I(Y^m;Z^m) + I(S;\theta_{\psi_S})]$. We previously stated $\mathcal{D}(Z^m)$ is achievable by imposing a constraint on the latent representation, $Z^m$ (see Corollary \ref{corr:bound}).

We now prove Corollary \ref{corr:bound} and derive $\mathcal{D}(Z^m)$. We consider the expansion of the term $I(X^m;Z^m)$:
\begin{align}
    \notag I(X^m;Z^m) &= H(Z^m) - H(Z^m | X^m) \\
    \notag &= H(Z_R^m + iZ_I^m) - H(g(X_R^m) + if(X_I^m) | X_R^m + iX_I^m) \\
    \notag &= H(Z_R^m) + H(iZ_I^m | Z_R^m) - H(Z_R^m | Z_R^m + iZ_I^m) \\
    \notag &= H(Z_R^m) + H(iZ_I^m | Z_R^m) \\ \label{eq:bound_I_X_Z}
    &\geq H(Z_R^m).
\end{align}
This lower bound is achievable when there exists a deterministic function, $\phi(\cdot)$, relating $Z_R^m$ and $Z_I^m$, wherein $H(iZ_I^m | Z_R^m) = 0$. We formalize this in Lemma \ref{lemma:I_X_Z}, where $\phi(\cdot)$ is bijective.
\begin{lemma} \label{lemma:I_X_Z}
Consider $X^m = X_R^m + iX_I^m \in \mathbb{C}^{dN}$, and $Z^m = Z_R^m + iZ_I^m \in \mathbb{C}^{lN}$. Subsequently, it follows that $I(X^m ; Z^m) \geq H(Z_R^m)$, with equality if the following condition holds $\forall m \in \mathcal{V}$:
\begin{align}
    \forall Z_I^m \in \mathbb{R}^{lN}, \ \exists! Z_R^m \in \mathbb{R}^{lN} : Z_I^m = \phi(Z_R^m) \implies I(X^m ; Z^m) = H(Z_R^m).
\end{align}
\end{lemma}
We now consider the expansion of $I(S;\theta_{\psi_S}) = H(S) - H(S | \theta_{\psi_S})$, and focus on the $H(S)$ term:
\begin{align}
    \notag H(S) &= H\Big((X^0,Y^0),(X^1,Y^1),\ldots,(X^{M-1},Y^{M-1})\Big) \\
    \notag &= \sum_{m=0}^{M-1} H \Big((X^m,Y^m) \Big| (X^{m-1}, Y^{m-1}), \ldots, (X^0, Y^0) \Big) \\ \label{eq:product_distribution}
    &= \sum_{m=0}^{M-1} H(X^m,Y^m) = M \big[ H(X^m,Y^m) \big].
\end{align}
We note Eq.~(\ref{eq:product_distribution}) follows from Section \ref{sec:preliminaries}, since $S \sim P^{\otimes M}$. Thus, $H(X^i, Y^i | X^j, Y^j) = H(X^i, Y^i)$, $\forall i, j \in \mathcal{V}$. Expanding $H(X^m,Y^m)$, we have that:
\begin{align}
    \notag H(S) &= M \big[H(X^m, Y^m, Z^m) - H(Z^m | X^m, Y^m) \big] \\
    \notag &= M \big[H(Y^m) + H(Z^m | Y^m) + H(X^m | Z^m, Y^m) \big] \\
    \notag &= M \big[ H(Y^m) + H(Z^m | Y^m) + H(X_R^m + iX_I^m | Z^m, Y^m) \big] \\
    \notag &= M \big[ H(Y^m) + H(Z^m | Y^m) + H(X_R^m | Z^m, Y^m) + H(X_R^m + iX_I^m | X_R^m, Z^m, Y^m) \\
    \notag &\quad \quad - H(X_R^m | X_R^m + iX_I^m, Z^m, Y^m) \big] \\
    \notag &= M \big[ H(Y^m) + H(Z^m, X_R^m | Y^m) + H(iX_I^m | X_R^m, Z^m, Y^m) \big] \\
    \notag &= M \big[ H(Y^m) + H(X_R^m | Y^m) + H(Z_R^m + iZ_I^m | X_R^m, Y^m) \\
    \notag &\quad \quad + H(iX_I^m | X_R^m, Z_R^m + iZ_I^m, Y^m) \big] \\
    \notag &= M \big[ H(Y^m) + H(X_R^m | Y^m) + H(iZ_I^m | X_R^m, Y^m) + H(iX_I^m | X_R^m, g(X_R^m) + iZ_I^m, Y^m) \big] \\ \label{eq:bound_H_Z_X_Y}
    &\geq M \big[H(Y^m) + H(X_R^m | Y^m) + H(iX_I^m | X_R^m, iZ_I^m, Y^m) \big].
\end{align}
As in Lemma \ref{lemma:I_X_Z}, this lower bound is achievable when there exists a deterministic function, $\phi(\cdot)$, relating $Z_R^m$ and $Z_I^m$, wherein we have $H(iZ_I^m | X_R^m, Y^m) = H(i\phi(f(X_R^m)) | X_R^m) = 0$. We formalize this in Lemma \ref{lemma:H_Z_X_Y}, where $\phi(\cdot)$ is bijective.
\begin{lemma} \label{lemma:H_Z_X_Y}
Consider $X^m = X_R^m + iX_I^m \in \mathbb{C}^{dN}$, with $Y^m \in \mathbb{C}^{k}$ and $Z^m = Z_R^m + iZ_I^m \in \mathbb{C}^{lN}$. We have that $H(S) \geq M [H(Y^m) + H(X_R^m | Y^m) + H(iX_I^m | X_R^m, iZ_I^m, Y^m)]$, with equality if the following condition holds $\forall m \in \mathcal{V}$:
\begin{equation}
\begin{gathered}
    \forall Z_I^m \in \mathbb{R}^{lN}, \ \exists! Z_R^m \in \mathbb{R}^{lN} : Z_I^m = \phi(Z_R^m) \implies \\
    H(S) = M \big[H(Y^m) + H(X_R^m | Y^m) + H(iX_I^m | X_R^m, iZ_I^m, Y^m) \big].
\end{gathered}
\end{equation}
\end{lemma}
Recalling the expansion of $I(S;\theta_{\psi_S}) = H(S) - H(S | \theta_{\psi_S})$, we now focus on the $-H(S | \theta_{\psi_S})$ term:
\begin{align}
    \notag -H(S | \theta_{\phi^S}) &= -H\Big((X^0,Y^0),(X^1,Y^1),\ldots,(X^{M-1},Y^{M-1})  \Big| \theta_{\psi_S}\Big) \\
    \notag &= -\sum_{m=0}^{M-1} H\Big(X^m,Y^m \Big| (X^{m-1}, Y^{m-1}),\ldots,(X^0, Y^0), \theta_{\psi_S}\Big) \\ \label{eq:product_distribution_cond}
    &= -\sum_{m=0}^{M-1} H(X^m,Y^m | \theta_{\psi_S}) = -M \big[H(X^m, Y^m | \theta_{\psi_S}) \big].
\end{align}
Paralleling Eq.~(\ref{eq:product_distribution}), we note that Eq.~(\ref{eq:product_distribution_cond}) also follows from Section \ref{sec:preliminaries}, since $S \sim P^{\otimes M}$. Accordingly, $H(X^i, Y^i | X^j, Y^j, \theta_{\psi_S}) = H(X^i, Y^i | \theta_{\psi_S})$, $\forall i, j \in \mathcal{V}$. Expanding $H(X^m, Y^m | \theta_{\psi_S})$, we have that:
\begin{align}
    \notag -H(S | \theta_{\psi_S}) &= -M\big[H(X^m | \theta_{\psi_S}) + H(Y^m | X^m, \theta_{\psi_S}) \big] \\
    \notag &= -M\big[H(X^m | \theta_{\psi_S}) + H(Y^m | X_R^m + iX_I^m, \theta_{f_S}, \theta_{g_S}) \big] \\
    \notag &= -M\big[H(X^m | \theta_{\psi_S}) - H(X_R^m + iX_I^m, \theta_{f_S}, \theta_{g_S}) + H(Y^m, X_R^m + iX_I^m, \theta_{f_S}, \theta_{g_S}) \big] \\ \label{eq:neg_H_S_theta}
    \notag &= -M\big[H(X^m | \theta_{\psi_S}) - H(X_R^m + iX_I^m, \theta_{g_S}) - H(\theta_{f_S} | X_R^m + iX_I^m, \theta_{g_S}) \\
    &\quad \quad \ \ \ + H(Y^m, X_R^m + iX_I^m, \theta_{g_S}) + H(\theta_{f_S} | Y^m, X_R^m + iX_I^m, \theta_{g_S}) \big].
\end{align}
We now consider the $H(\theta_{f_S} | Y^m, X_R^m + iX_I^m, \theta_{g_S}) - H(\theta_{f_S} | X_R^m + iX_I^m, \theta_{g_S})$ term. By the properties of conditional entropy, we observe that:
\begin{align} \label{eq:cond_entropy}
     H(\theta_{f_S} | Y^m, X_R^m + iX_I^m, \theta_{g_S}) - H(\theta_{f_S} | X_R^m + iX_I^m, \theta_{g_S}) \leq 0.
\end{align}
Where equality follows if $H(\theta_{f_S} | X_R^m + iX_I^m, \theta_{g_S}) = H(\theta_{f_S} | Y^m, X_R^m + iX_I^m, \theta_{g_S})$. We now show that this condition is met when $\theta_{f_S}$ is deterministic given $\theta_{g_S}$ and $X_R^m + iX_I^m$.

Suppose we are given $X_R^m + iX_I^m$ and $\theta_{g_S}$. It follows that $Z_R^m = g_S(X_R^m)$ is deterministic. We now impose the constraint presented in Lemma \ref{lemma:I_X_Z} and \ref{lemma:H_Z_X_Y}, wherein there exists a bijective, deterministic function, $\phi(\cdot)$, such that $Z_I^m = \phi(Z_R^m), \, \forall m \in \mathcal{V}$. It follows that $Z_I^m = \phi(g_S(X_R^m))$ is deterministic. Per Eq. (\ref{eq:cond_entropy}), the following equalities now hold under the imposed constraint:
\begin{align}
    Z_I^m = \phi(Z_R^m) \implies H(\theta_{f_S} | X_R^m + iX_I^m, \theta_{g_S}) &= H(\theta_{f_S} | X_R^m + iX_I^m, Z_I^m), \\
    Z_I^m = \phi(Z_R^m) \implies H(\theta_{f_S} | Y^m, X_R^m + iX_I^m, \theta_{g_S}) &= H(\theta_{f_S} | Y^m, X_R^m + iX_I^m, Z_I^m).
\end{align}
We also recall the Markov chain presented in Figure \ref{fig:markov_steinmetz}. The Steinmetz network architecture informs us that $Y^m$ does not reduce the uncertainty in $\theta_{f_S}$ given $X_I^m$ and $Z_I^m$. Therefore, we have that:
\begin{align}
    Z_I^m = \phi(Z_R^m) &\implies H(\theta_{f_S} | Y^m, X_R^m + iX_I^m, \theta_{g_S}) - H(\theta_{f_S} | X_R^m + iX_I^m, \theta_{g_S}) = 0, \\
    \notag -H(S | \theta_{\psi_S}) &\geq -M \big[H(X^m | \theta_{\psi_S}) - H(X_R^m + iX_I^m, \theta_{g_S}) + H(Y^m, X_R^m + iX_I^m, \theta_{g_S}) \big] \\ \label{eq:bound_H_S_theta}
    &= -M \big[H(X^m | \theta_{\psi_S}) - H(Y^m | X_R^m + iX_I^m, \theta_{g_S}) \big].
\end{align}
We summarize the achievability of this lower bound in Lemma \ref{lemma:-H_S_theta}.
\begin{lemma} \label{lemma:-H_S_theta}
Consider $X^m = X_R^m + iX_I^m \in \mathbb{C}^{dN}$, $Y^m \in \mathbb{C}^{k}$ and $Z^m = Z_R^m + iZ_I^m \in \mathbb{C}^{lN}$, with $\theta_{\psi_S} \in \mathbb{R}^c, \theta_{g_S} \in \mathbb{R}^{c_2}$. We have that $-H(S | \theta_{\psi_S}) \geq -M[H(X^m | \theta_{\psi_S}) - H(Y^m | X_R^m + iX_I^m, \theta_{g_S})]$, with equality if the following condition holds $\forall m \in \mathcal{V}$:
\begin{equation}
\begin{gathered}
    \forall Z_I^m \in \mathbb{R}^{lN}, \ \exists! Z_R^m \in \mathbb{R}^{lN} : Z_I^m = \phi(Z_R^m) \implies \\
    -H(S | \theta_{\psi_S}) = -M \big[ H(X^m | \theta_{\psi_S}) - H(Y^m | X_R^m + iX_I^m, \theta_{g_S}) \big].
\end{gathered}
\end{equation}
\end{lemma}
We can now determine the overall lower bound, $\mathcal{D}(Z^m)$, on $[I(X^m;Z^m) - I(Y^m;Z^m) + I(S;\theta_{\psi_S})]$ by substituting Eq. (\ref{eq:bound_I_X_Z}), (\ref{eq:bound_H_Z_X_Y}), and (\ref{eq:bound_H_S_theta}) into Eq. (\ref{eq:conjecture_bound}):
\begin{align}
    &\mathcal{D}(Z^m) \leq I(X^m;Z^m) - I(Y^m;Z^m) + I(S;\theta_{\psi_S}), \\
    \notag &\text{where: } \mathcal{D}(Z^m) = H(Z_R^m) - I(Y^m ; Z^m) - M \big[H(X^m | \theta_{\psi_S}) - H(Y^m | X_R^m + iX_I^m, \theta_{g_S}) \\
    \notag &\qquad \qquad \qquad \quad \quad \ - H(Y^m) - H(X_R^m | Y^m) - H(iX_I^m | X_R^m, iZ_I^m, Y^m) \big].
\end{align}
Revisiting Corollary \ref{corr:bound}, we note that $\mathcal{D}(Z^m)$ is an achievable lower bound, with equality observed when the imposed condition, $f \in \mathcal{F}^m$, delineates $Z_I^m$ and $Z_R^m$ as being related by a deterministic, bijective function, $\phi(\cdot)$. We summarize this result in Lemma \ref{lemma:overall_lower_bound}.
\begin{lemma} \label{lemma:overall_lower_bound}
Consider $X^m = X_R^m + iX_I^m \in \mathbb{C}^{dN}$, $Y^m \in \mathbb{C}^{k}$, and $Z^m = Z_R^m + iZ_I^m \in \mathbb{C}^{lN}$ with $\theta_{\psi_S} \in \mathbb{R}^c, \theta_{g_S} \in \mathbb{R}^{c_2}$. It follows that $\mathcal{D}(Z^m) \leq I(X^m;Z^m) - I(Y^m;Z^m) + I(S;\theta_{\psi_S})$, where:
\begin{align}
\begin{split}
    \mathcal{D}(Z^m) &= H(Z_R^m) - I(Y^m ; Z^m) - M \big[H(X^m | \theta_{\psi_S}) - H(Y^m | X_R^m + iX_I^m, \theta_{g_S}) \\
    &\quad \quad - H(Y^m) - H(X_R^m | Y^m) - H(iX_I^m | X_R^m, iZ_I^m, Y^m) \big].
\end{split}
\end{align}
With equality if the following condition holds $\forall m \in \mathcal{V}$:
\begin{align}
\begin{gathered}
    \forall Z_I^m \in \mathbb{R}^{lN}, \ \exists! Z_R^m \in \mathbb{R}^{lN} : Z_I^m = \phi(Z_R^m) \implies \\
    I(X^m;Z^m) - I(Y^m;Z^m) + I(S;\theta_{\psi_S}) = \mathcal{D}(Z^m).
\end{gathered}
\end{align}
\end{lemma}
This result is also summarized in Theorem \ref{theorem:overall_lower_bound} of the main text. We extend this result to derive the smaller upper bound on the generalization gap, $\Delta s$, provided in Theorem \ref{theorem:generalization_error_smaller} of the main text.

\section{Additional Proofs} 

\subsection{Complementarity Principle} \label{sec:complementarity_principle}
For completeness, we prove Corollary \ref{corr:complementarity} from the main text. Per the notation outlined in Section \ref{sec:steinmetz_theory}, we first note the expansions of the auto-covariance matrices, $\mathbf{K}_{X_R^m}$ and $\mathbf{K}_{X_I^m}$:
\begin{align}
    \begin{split}
    \mathbf{K}_{X_R^m} &= 
    \left[ \begin{array}{cc}
        \mathbf{K}_{Z_R^m} & \mathbf{K}_{Z_R^m, \Lambda_R^m} \\
        \mathbf{K}_{\Lambda_R^m, Z_R^m} & \mathbf{K}_{\Lambda_R^m} \\
    \end{array} \right] = 
    \left[ \begin{array}{cc}
        \mathbf{K}_{Z_R^m} & \mathbf{0}_{kN \times (d-k)N} \\
        \mathbf{0}_{(d-k)N \times kN} & \mathbf{K}_{\Lambda_R^m} \\
    \end{array} \right]
    \end{split}, \\
    \begin{split}
    \mathbf{K}_{X_I^m} &= 
    \left[ \begin{array}{cc}
        \mathbf{K}_{Z_I^m} & \mathbf{K}_{Z_I^m, \Lambda_I^m} \\
        \mathbf{K}_{\Lambda_I^m, Z_I^m} & \mathbf{K}_{\Lambda_I^m} \\
    \end{array} \right] = 
    \left[ \begin{array}{cc}
        \mathbf{K}_{Z_I^m} & \mathbf{0}_{kN \times (d-k)N} \\
        \mathbf{0}_{(d-k)N \times kN} & \mathbf{K}_{\Lambda_I^m} \\
    \end{array} \right]
    \end{split}.
\end{align}
Regarding the cross-covariance matrices, $\mathbf{K}_{X_I^m, X_R^m} = \mathbf{K}_{X_R^m, X_I^m}^T$ and $\overline{\mathbf{K}}_{X_I^m, X_R^m} = \overline{\mathbf{K}}\vphantom{\mathbf{K}}_{X_R^m, X_I^m}^T$, where:
\begin{align}
    \begin{split}
    \mathbf{K}_{X_R^m, X_I^m} &= 
    \left[ \begin{array}{cc}
        \mathbf{K}_{Z_R^m, Z_I^m} & \mathbf{K}_{Z_R^m, \Lambda_I^m} \\
        \mathbf{K}_{\Lambda_R^m, Z_I^m} & \mathbf{K}_{\Lambda_R^m, \Lambda_I^m} \\
    \end{array} \right], \quad  
    \overline{\mathbf{K}}_{X_R^m, X_I^m} = 
    \left[ \begin{array}{cc}
        \mathbf{0}_{kN \times kN} & \mathbf{K}_{Z_R^m, \Lambda_I^m} \\
        \mathbf{K}_{\Lambda_R^m, Z_I^m} & \mathbf{K}_{\Lambda_R^m, \Lambda_I^m} \\
    \end{array} \right].
    \end{split}
\end{align}
We now derive and compare the $L_{p,q}$ norm of $\mathbf{\Sigma_J}$ and $\mathbf{\Sigma_S}$. We first note that $\mathbf{K}_{Z_R^m,Z_I^m} = \mathbf{0}_{kN \times kN}$ for the separate-then-joint processing case, since $\smash{f(\cdot)}$ and $\smash{g(\cdot)}$ do not consider the interactions between $Z_R^m$ and $Z_I^m$. These interactions are not lost, however, as they are leveraged by $h^*(\cdot)$ during the joint processing step. Accordingly, it follows that:
\begin{align}
    \notag \|\mathbf{\Sigma_J}\|_{p,q} &= \left(\sum_{i=1}^{dN} \left(\|[\mathbf{K}_{X_R^m}]_i\|_q^q + \|[\mathbf{K}_{X_R^m, X_I^m}]_i\|_q^q \right)^{\frac{p}{q}} + \sum_{i=1}^{dN} \left(\|[\mathbf{K}_{X_I^m, X_R^m}]_i\|_q^q + \|[\mathbf{K}_{X_I^m}]_i\|_q^q\right)^{\frac{p}{q}}\right)^{\frac{1}{p}} \\
    \notag &= \Bigg( \sum_{i=1}^{kN} \left(\|[\mathbf{K}_{Z_R^m}]_i\|_q^q + \|[\mathbf{K}_{Z_R^m, Z_I^m}]_i\|_q^q + \|[\mathbf{K}_{Z_R^m, \Lambda_I^m}]_i\|_q^q \right)^{\frac{p}{q}} \\
    \notag &\quad \quad \quad + \sum_{i=1}^{(d-k)N} \left(\|[\mathbf{K}_{\Lambda_R^m}]_i \|_q^q + \|[\mathbf{K}_{\Lambda_R^m, Z_I^m}]_i\|_q^q + \|[\mathbf{K}_{\Lambda_R^m, \Lambda_I^m}]_i\|_q^q \right)^{\frac{p}{q}} \\
    \notag &\quad \quad \quad + \sum_{i=1}^{kN} \left(\|[\mathbf{K}_{Z_I^m}]_i\|_q^q + \|[\mathbf{K}_{Z_I^m, Z_R^m}]_i\|_q^q + \|[\mathbf{K}_{Z_I^m, \Lambda_R^m}]_i\|_q^q \right)^{\frac{p}{q}} \\
    \notag &\quad \quad \quad + \sum_{i=1}^{(d-k)N} \left(\|[\mathbf{K}_{\Lambda_I^m}]_i \|_q^q + \|[\mathbf{K}_{\Lambda_I^m, Z_R^m}]_i\|_q^q + \|[\mathbf{K}_{\Lambda_I^m, \Lambda_R^m}]_i\|_q^q \right)^{\frac{p}{q}} \Bigg)^{\frac{1}{p}} \\
    \notag &\geq \Bigg( \sum_{i=1}^{kN} \left(\|[\mathbf{K}_{Z_R^m}]_i\|_q^q + \|[\mathbf{K}_{Z_R^m, \Lambda_I^m}]_i\|_q^q \right)^{\frac{p}{q}} \\
    \notag &\quad \quad \quad + \sum_{i=1}^{(d-k)N} \left(\|[\mathbf{K}_{\Lambda_R^m}]_i \|_q^q + \|[\mathbf{K}_{\Lambda_R^m, Z_I^m}]_i\|_q^q + \|[\mathbf{K}_{\Lambda_R^m, \Lambda_I^m}]_i\|_q^q \right)^{\frac{p}{q}} \\
    \notag &\quad \quad \quad + \sum_{i=1}^{kN} \left(\|[\mathbf{K}_{Z_I^m}]_i\|_q^q + \|[\mathbf{K}_{Z_I^m, \Lambda_R^m}]_i\|_q^q \right)^{\frac{p}{q}} \\
    \notag &\quad \quad \quad + \sum_{i=1}^{(d-k)N} \left(\|[\mathbf{K}_{\Lambda_I^m}]_i \|_q^q + \|[\mathbf{K}_{\Lambda_I^m, Z_R^m}]_i\|_q^q + \|[\mathbf{K}_{\Lambda_I^m, \Lambda_R^m}]_i\|_q^q \right)^{\frac{p}{q}} \Bigg)^{\frac{1}{p}} \\
    \notag &= \Bigg(\sum_{i=1}^{dN} \left(\|[\mathbf{K}_{X_R^m}]_i\|_q^q + \|[\overline{\mathbf{K}}_{X_R^m, X_I^m}]_i\|_q^q \right)^{\frac{p}{q}} + \sum_{i=1}^{dN} \left(\|[\overline{\mathbf{K}}_{X_I^m, X_R^m}]_i\|_q^q + \|[\mathbf{K}_{X_I^m}]_i\|_q^q\right)^{\frac{p}{q}}\Bigg)^{\frac{1}{p}} \\
    \notag &= \|\mathbf{\Sigma_S}\|_{p,q}
\end{align}
Therefore, $\|\mathbf{\Sigma_J}\|_{p,q} \geq \|\mathbf{\Sigma_S}\|_{p,q}$. We also note $\mathbf{K}_{Z_R^m, Z_I^m} = \mathbf{K}_{Z_I^m, Z_R^m} = \mathbf{0}_{dN \times dN}$ when $Z_R^m \ind Z_I^m$. Accordingly, it follows that $Z_R^m \ind Z_I^m \implies \|\mathbf{\Sigma_J}\|_{p,q} = \|\mathbf{\Sigma_S}\|_{p,q}$.

\subsection{Orthogonality of Latent Analytic Signal Representation} \label{sec:hilbert_orthogonality}
We now prove Corollary \ref{corr:hilbert_constraint} from the main text, which claims $Z_I^m = \mathcal{H}\{Z_R^m \}$ enforces orthogonality between $Z_R^m$ and $Z_I^m$. Let $\mathbf{F}_R^m = \mathcal{F}\{ Z_R^m \} \in \mathbb{C}^{lN}$ denote the DFT of $Z_R^m$ and let $\mathbf{H}_R^m \in \mathbb{C}^{lN}$ denote the frequency components of $\mathcal{H}\{Z_R^m\}$. We consider the inner product $\langle \cdot, \cdot \rangle : \mathcal{Z} \times \mathcal{Z} \rightarrow \mathbb{R}_{\geq 0}$, wherein:
\begin{align}
    \notag \langle Z_R^m, Z_I^m \rangle &= \mathbb{E}[Z_R^m Z_I^m] = \mathbb{E}[Z_R^m \ \mathcal{H}\{ Z_R^m\}] = \mathbb{E}\left[ \sum_{n = 0}^{lN-1} Z_R^m[n] \mathcal{H}\{ Z_R^m\}[n]\right] \\
    &= \mathbb{E}\left[ \sum_{n = 0}^{lN-1} Z_R^m[n] \left(\frac{1}{lN} \sum_{b=0}^{lN-1} \mathbf{H}_R^m[b] e^{\frac{i2\pi bn}{lN}}\right)\right]
\end{align} 
We now substitute Eq.~(\ref{eq:phase_shift}) into the above expression, and expand the $\mathbf{H}_R^m[b]$ term.
\begin{align}
\begin{split}
    \langle Z_R^m, Z_I^m \rangle &= \mathbb{E}\left[ \sum_{n = 0}^{lN-1} Z_R^m[n] \left(\frac{1}{lN} \sum_{b=0}^{lN-1} \mathbf{F}_R^m[b] \cdot (\pm i) e^{\frac{i2\pi bn}{lN}}\right)\right] \\
    &= \mathbb{E}\left[\frac{\pm i}{lN} \sum_{b=0}^{lN-1} \mathbf{F}_R^m[b] \sum_{n = 0}^{lN-1} Z_R^m[n] e^{\frac{i2\pi bn}{lN}}\right] \\
    &= \mathbb{E}\left[ \frac{\pm i}{lN} \sum_{b=0}^{lN-1} \mathbf{F}_R^m[b] \mathbf{F}_R^m[b]^*\right] = \mathbb{E}\left[ \frac{\pm i}{lN} \sum_{b=0}^{lN-1} | \mathbf{F}_R^m[b]|^2 \text{sgn}(b) \right] \\
    &= 0.
\end{split}
\end{align}
Since $\text{sgn}(b)$ is odd and $|\mathbf{F}_R^m[b]|^2$ is even, the sum evaluates to zero. Therefore, when $Z_I^m = \mathcal{H}\{Z_R^m \}$, $\langle Z_R^m, Z_I^m \rangle = 0$, and consequently, $Z_R^m$ and $Z_I^m$ are orthogonal.

\section{Dataset Descriptions} \label{sec:dataset_descriptions}

\subsection{CV-MNIST Dataset}
The \textbf{MNIST} dataset is a collection of handwritten digits commonly used to train image processing systems. The CV-MNIST dataset is formed by taking a $dN = 784$-point DFT of each flattened MNIST image, yielding a $728$-dimensional real feature vector, and a $728$-dimensional imaginary feature vector. For the classification result in Section \ref{sec:benchmark_datasets}, we consider the first $M = 500$ training samples from CV-MNIST and for the noise robustness result, we consider $M = 60{,}000$ training samples from CV-MNIST. We consider $10{,}000$ test samples in both cases. The features and labels within CV-MNIST are summarized as follows:
\begin{itemize}
    \item Each feature (image) is represented as a 728-dimensional real vector and a 728-dimensional imaginary vector, obtained by taking the DFT of the original size $28 \times 28$ grayscale image.
    \item Target Variable: The numerical class (digit) the image represents, ranging from 1 to 10 ($k = 10$).
\end{itemize}

\subsection{CV-CIFAR-10 Dataset}
The \textbf{CIFAR-10} dataset is a collection of color images categorized into $10$ different classes, and is commonly used to train image processing systems. The CV-CIFAR-10 dataset is formed by taking a $dN = 3072$-point DFT of each flattened CIFAR-10 image, yielding a $3072$-dimensional real feature vector, and a $3072$-dimensional imaginary feature vector. For the classification and noise robustness results from Section \ref{sec:benchmark_datasets}, we consider $M = 50{,}000$ training samples and $10{,}000$ test samples from CV-CIFAR-10. The features and labels within CV-CIFAR-10 are summarized as follows:
\begin{itemize}
    \item Each feature (image) is represented as a 3072-dimensional real vector and a 3072-dimensional imaginary vector, obtained by taking the DFT of the original size $32 \times 32 \times 3$ image.
    \item Target Variable: The numerical class (category) the image represents, ranging from 1 to 10 ($k = 10$).
\end{itemize}

\subsection{CV-CIFAR-100 Dataset}
The \textbf{CIFAR-100} dataset is a collection of color images categorized into $100$ different classes, and is commonly used to train image processing systems. The CV-CIFAR-100 dataset is formed by taking a $dN = 3072$-point DFT of each flattened CIFAR-100 image, yielding a $3072$-dimensional real feature vector and a $3072$-dimensional imaginary feature vector. For the classification result from Section \ref{sec:benchmark_datasets}, we consider $M = 50{,}000$ training samples and $10{,}000$ test samples from CV-CIFAR-100. The features and labels within CV-CIFAR-100 are summarized as follows:
\begin{itemize} 
    \item Each feature (image) is represented as a 3072-dimensional real vector and a 3072-dimensional imaginary vector, obtained by taking the DFT of the original size $32 \times 32 \times 3$ image. 
    \item Target Variable: The numerical class (category) the image represents, ranging from 1 to 100 ($k = 100$).
\end{itemize}

\subsection{CV-FSDD Dataset}
The \textbf{FSDD} (Free Spoken Digit Dataset) is a collection of audio recordings of spoken digits categorized into $10$ different classes, and is commonly used to train audio processing systems. The CV-FSDD dataset is formed by taking a $dN = 8000$-point DFT of each flattened FSDD recording, yielding a $8000$-dimensional real feature vector and a $8000$-dimensional imaginary feature vector. For the classification result from Section \ref{sec:benchmark_datasets}, we consider $M = 2{,}700$ training samples and $300$ test samples from CV-FSDD. The features and labels within CV-FSDD are summarized as follows:
\begin{itemize}
    \item Each feature (audio recording) is represented as a 8000-dimensional real vector and a 8000-dimensional imaginary vector, obtained by taking the DFT of the original audio signal.
    \item Target Variable: The numerical digit (category) the recording represents, ranging from 1 to 10 ($k = 10$).
\end{itemize}

\subsection{RASPNet Dataset}
The \textbf{RASPNet} dataset for radar adaptive signal processing consists of radar returns gathered from $100$ realistic radar scenarios located in the contiguous United States. We consider the scenario index $i = 29$ from RASPNet, which corresponds to the Bonneville Salt Flats, UT, and consider the real and imaginary parts of a size $5 \times 21 \times 16$ radar return, comprising $5$ realizations, $21$ range bins, and $16$ channels (single pulse transmission). Accordingly, each sample is defined by a $dN = 1680$-dimensional real feature vector, and a $1680$-dimensional imaginary feature vector. Each radar return contains a single point target, with position encoded in Cartesian coordinates (x, y). For the regression result from Section \ref{sec:benchmark_datasets}, we consider $M = 20{,}000$ training samples and $5{,}000$ test samples from scenario $i = 29$, which can be accessed \href{https://shyamven.github.io/RASPNet/download.html}{here}. The features and labels are summarized as follows:
\begin{itemize}
    \item Each feature (radar return) posits a 1680-dimensional real vector and a 1680-dimensional imaginary vector.
    \item Target Variable: The target location in Cartesian coordinates (x, y) ($k = 2$).
\end{itemize}

\section{Neural Network Architectures} \label{sec:network_architectures}
We provide a detailed description of three different neural network architectures designed for classification and regression. Each of these architectures were employed to generate the respective empirical results pertaining to the aforementioned tasks.

\subsection{Steinmetz Neural Network}
The Steinmetz Network is designed to handle both real and imaginary components of the input data separately before combining them for the final prediction. This architecture can be applied to both classification and regression tasks (see Figure \ref{fig:real_architectures}).

\begin{itemize}
    \item \textbf{Fully Connected Layer (\texttt{realfc1})}: Transforms the real part of the input features to a higher dimensional space. It takes $dN$-dimensional inputs and yields $lN$-dimensional outputs.
    \item \textbf{ReLU Activation (\texttt{realrelu1})}: Introduces non-linearity to the model. It operates element-wise on the output of \texttt{realfc1}.
    \item \textbf{Fully Connected Layer (\texttt{realfc2})}: Further processes the output of \texttt{realrelu1}, yielding $lN$-dimensional outputs.
    \item \textbf{ReLU Activation (\texttt{realrelu2})}: Applies non-linearity to the output of \texttt{realfc2}.
    \item \textbf{Fully Connected Layer (\texttt{imagfc1})}: Transforms the imaginary part of the input features to a higher dimensional space, paralleling \texttt{realfc1}.
    \item \textbf{ReLU Activation (\texttt{imagrelu1})}: Applies non-linearity to the output of \texttt{imagfc1}.
    \item \textbf{Fully Connected Layer (\texttt{imagfc2})}: Further processes the output of \texttt{imagrelu1}, yielding $lN$-dimensional outputs.
    \item \textbf{ReLU Activation (\texttt{imagrelu2})}: Applies non-linearity to the output of \texttt{imagfc2}.
    \item \textbf{Fully Connected Layer (\texttt{regressor})}: Combines the extracted features from both networks into a single $2lN$-dimensional feature vector (the latent space), which is then passed through a fully connected layer to produce the final output of dimension $k$.
\end{itemize}

\subsection{Real-Valued Neural Network}
\begin{figure*}[t!]
    \centering
    \begin{subfigure}{0.4\textwidth}
    \includegraphics[width=\textwidth]{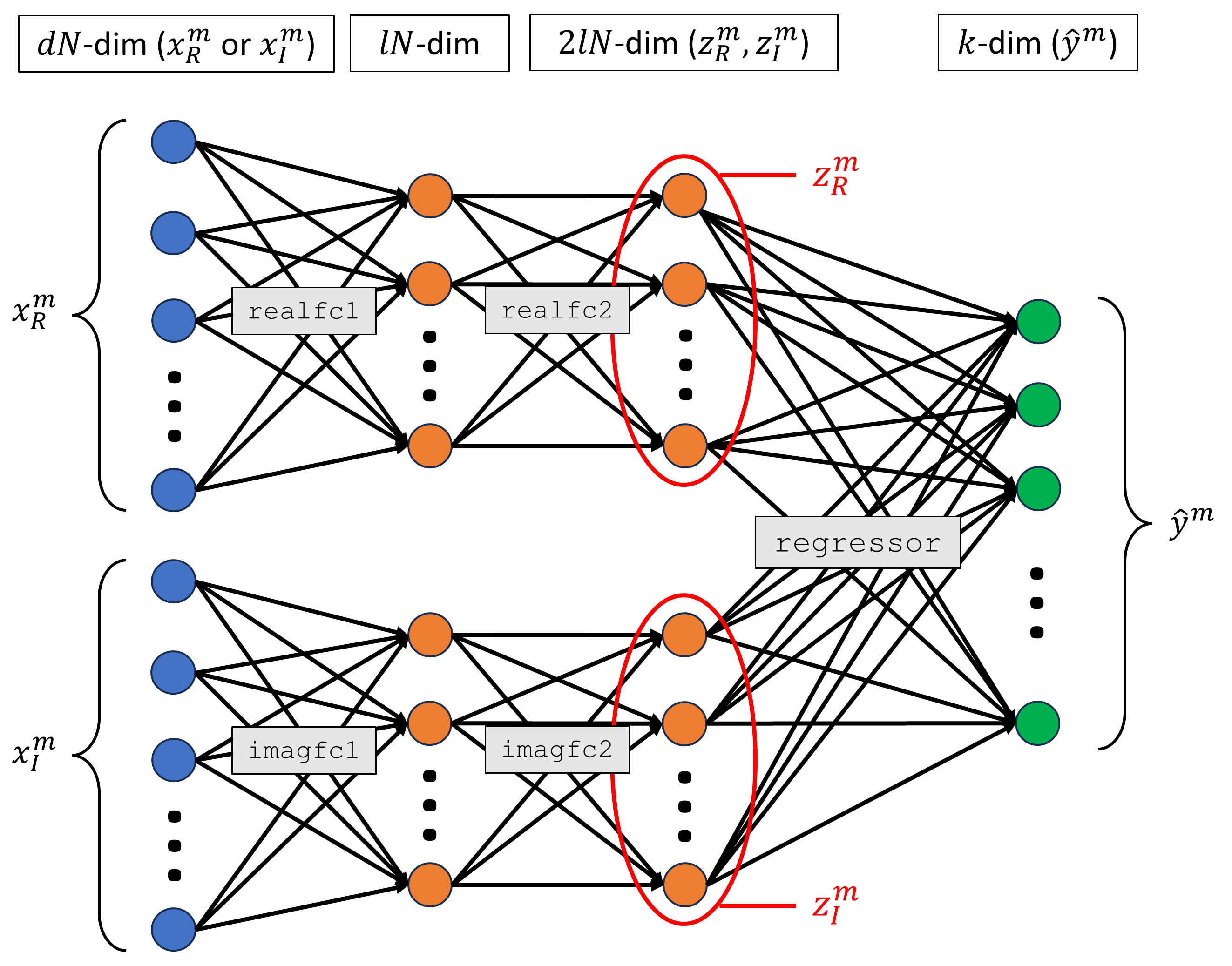}
    \caption{Steinmetz neural network architecture}
    \end{subfigure}
    \begin{subfigure}{0.4\textwidth}
        \includegraphics[width=\textwidth]{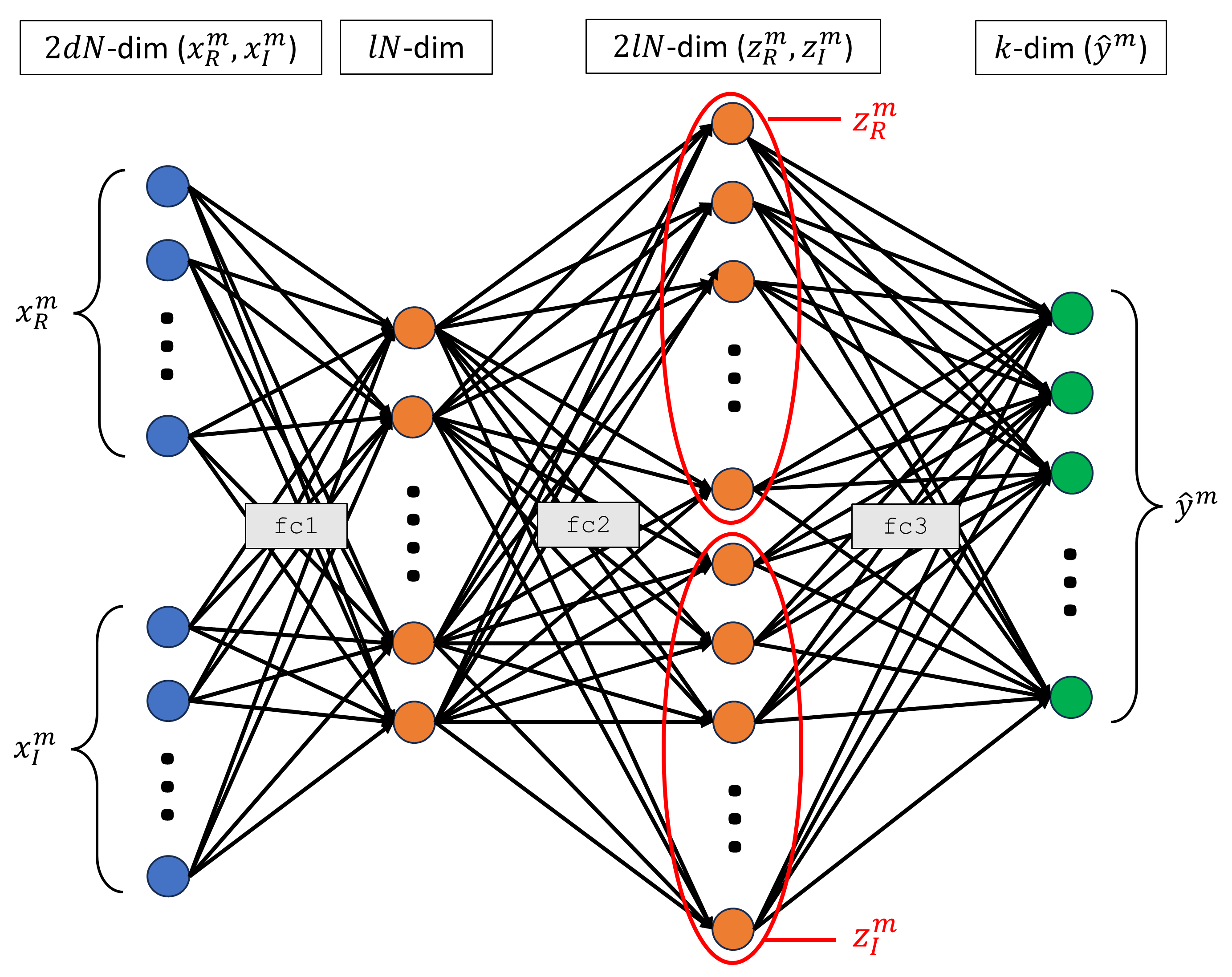}
        \caption{Real-valued neural network architecture}
    \end{subfigure}
    \caption{Real-valued architectures for complex-valued data processing.}
    \label{fig:real_architectures}
\end{figure*}

The Real-valued neural network (RVNN) architecture is a straightforward and effective approach for handling both real and imaginary components by concatenating them and processing them together. It can be used for both classification and regression tasks (see Figure \ref{fig:real_architectures}).

\begin{itemize}
\item \textbf{Fully Connected Layer (\texttt{fc1})}: Takes the concatenated real and imaginary components ($2dN$-dimensional) as input and produces $lN$-dimensional features.
\item \textbf{ReLU Activation (\texttt{relu1})}: Introduces non-linearity to the model after \texttt{fc1}.
\item \textbf{Fully Connected Layer (\texttt{fc2})}: Further processes the output of \texttt{relu1}, yielding the $2lN$-dimensional latent space as the output.
\item \textbf{ReLU Activation (\texttt{relu2})}: Applies non-linearity to the output of \texttt{fc2}.
\item \textbf{Fully Connected Layer (\texttt{fc3})}: Produces the final output of dimension $k$.
\end{itemize}

\subsubsection{Complex-Valued Neural Network}
\begin{figure*}[t!]
\centering
\includegraphics[width=0.42\linewidth]{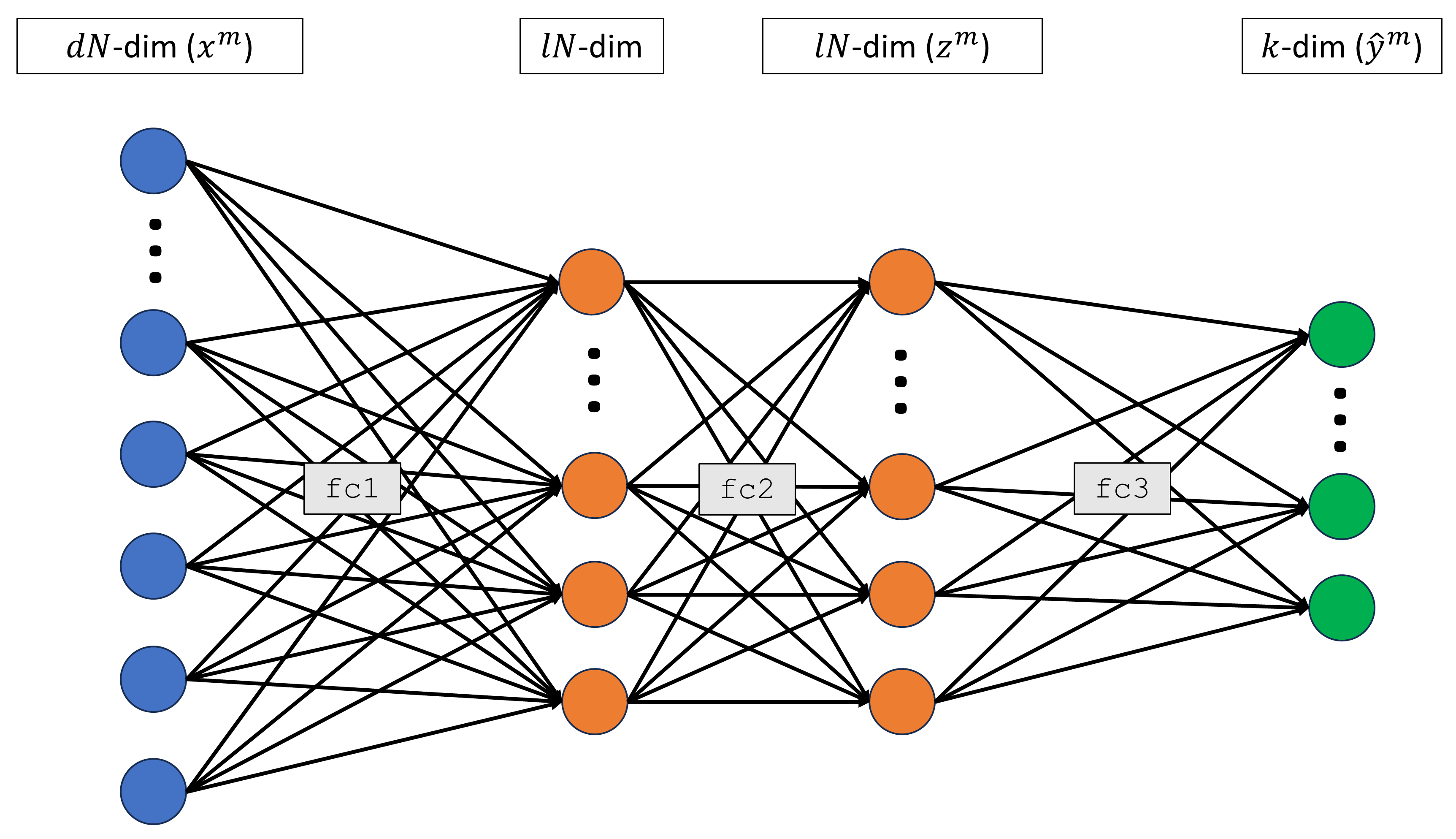}
\caption{Complex-valued neural network architecture.}
\label{fig:cvnn_architecture}
\end{figure*}

The Complex-Valued Neural Network (CVNN) is designed to handle complex-valued data by treating the real and imaginary parts jointly as complex numbers. It can be used in classification and regression tasks (see Figure \ref{fig:cvnn_architecture}). For classification tasks, we take the magnitude of the \texttt{fc3} layer output.

\begin{itemize}
    \item \textbf{Complex Linear Layer (\texttt{fc1})}: Transforms the $dN$-dimensional complex inputs into $lN$-dimensional complex features.
    \item \textbf{Complex ReLU Activation (\texttt{relu1})}: Applies complex-valued ReLU activation after \texttt{fc1}.
    \item \textbf{Complex Linear Layer (\texttt{fc2})}: Further processes the $lN$-dimensional complex features into $lN$-dimensional complex features (the latent space).
    \item \textbf{Complex ReLU Activation (\texttt{relu2})}: Applies complex-valued ReLU activation after \texttt{fc2}.
    \item \textbf{Complex Linear Layer (\texttt{fc3})}: Produces the final $k$-dimensional output by transforming the $lN$-dimensional complex features.
\end{itemize}

\newpage

\subsection{Neural Network Training Hyperparameters}
The relevant hyperparameters used to train the neural networks from Appendix Section \ref{sec:network_architectures} are provided in Table \ref{tab:hyperparameters_table}. All results presented in the main text were produced using these hyperparameter choices.
\begin{table*}[h]
\caption{Neural network training hyperparameters (grouped by dataset/task).}
\label{tab:hyperparameters_table}
\begin{center}
\begin{tabular}{c|c|c|c|c}
\hline 
\textbf{Dataset/Task} & \textbf{Experiment} & \textbf{Optimizer} & \thead{\textbf{Learning} \\ \textbf{Rate} ($\boldsymbol{\alpha}$)} & \thead{\textbf{Consistency penalty} ($\boldsymbol{\beta}$) \\ (analytic neural network)} \\
\hline
CV-MNIST & No ablations & Adam & 0.001 & 0.001 \\
CV-MNIST & Additive noise & Adam & 0.001 & 0.001 \\
\hline
CV-CIFAR-10 & No ablations & Adam & 0.0001 & 0.001 \\
CV-CIFAR-10 & Additive noise & Adam & 0.0001 & 0.001 \\
\hline
CV-CIFAR-100 & No ablations & Adam & 0.0001 & 0.001 \\
\hline
CV-FSDD & No ablations & Adam & 0.0002 & 1.0\text{e-}6 \\
\hline
RASPNet & No ablations & Adam & 0.005 & 0.001 \\
\hline
Channel Identification & SNR = $5$ dB & Adam & 0.0001 & 0.0001 \\
\hline
\end{tabular}
\end{center}
\end{table*}

\end{document}